%% file: arxiv.tex
\documentclass{article}
\usepackage[margin=1in]{geometry}

\usepackage{amssymb,amsmath,amsthm,bbm,mathtools}
\usepackage{float}
\usepackage{algorithm}
\usepackage[noend]{algorithmic}
\input{math}

\usepackage[utf8]{inputenc} 
\usepackage[T1]{fontenc}    
\usepackage{booktabs}       
\usepackage{amsfonts}       
\usepackage{nicefrac}       
\usepackage{microtype}      
\usepackage{xcolor}         
\usepackage{graphicx}
\usepackage{subcaption}
\usepackage{multirow}
\usepackage[shortlabels]{enumitem}
\usepackage[normalem]{ulem}

\usepackage[colorlinks=true,citecolor=blue,urlcolor=blue,linkcolor=black]{hyperref}

\usepackage{natbib}
\setcitestyle{authoryear,open={(},close={)}}

\newcommand{\mc}{\text{MC}}
\newcommand{\ma}{\text{MA}}
\newcommand{\nma}{\textnormal{MA}}

\newcommand{\pred}{\text{pred}}
\newcommand{\npredt}{\textnormal{pred}}

\newif\ifcomments
\commentstrue

\ifcomments
    \newcommand{\jivat}[1]{\textcolor{olive}{#1 --Jivat}}
    \newcommand{\isaac}[1]{\textcolor{cyan}{#1 --Isaac}}
    \newcommand{\removable}[1]{\textcolor{brown}{#1}}
\else
    \newcommand{\jivat}[1]{}
    \newcommand{\isaac}[1]{}
    \newcommand{\removable}[1]{\textcolor{brown}{}}
\fi

\makeatletter
\newcommand\footnoteref[1]{\protected@xdef\@thefnmark{\ref{#1}}\@footnotemark}
\makeatother

\title{Locally Adaptive Multi-Objective Learning}

\author{Jivat Neet Kaur$^*$ \and  Isaac Gibbs$^*$ \and
Michael I.\ Jordan$^{*\dagger}$} 
\date{$^*$University of California, Berkeley \qquad $^\dagger$Inria, Paris} 

\begin{document}

\maketitle

\begin{abstract}
We consider the general problem of learning a predictor that satisfies multiple objectives of interest simultaneously, a broad framework that captures a range of specific learning goals including calibration, regret, and multiaccuracy.~We work in an online setting where the data distribution can change arbitrarily over time. Existing approaches to this problem aim to minimize the set of objectives over the \textit{entire time horizon} in a worst-case sense, and in practice they do not necessarily adapt to distribution shifts. Earlier work has aimed to alleviate this problem by incorporating additional objectives that target local guarantees over contiguous subintervals.~Empirical~evaluation of these proposals is, however, scarce.~In~this~article,~we consider an alternative procedure that achieves local adaptivity by replacing one part of the multi-objective learning method with an adaptive online algorithm.~Empirical~evaluations~on datasets~from~energy forecasting and algorithmic fairness show that~our~proposed~method improves upon existing approaches and achieves unbiased predictions over subgroups, while remaining robust under distribution shift.
\end{abstract}

\input{files/1_intro}

\input{files/2_method}
\input{files/5_expts}

\input{files/6_extensions}

\input{files/ack}

\bibliography{ref}
\bibliographystyle{plainnat}

\newpage 
\appendix

\input{files/app_proofs}

\input{files/app_expt_details}
\input{files/app_further_expts}

\end{document}

%% file: math.tex
\newtheorem{theorem}{Theorem}
\newtheorem{corollary}{Corollary}
\newtheorem{lemma}{Lemma}

\newtheorem{assumption}{Assumption}
\theoremstyle{definition}
\newtheorem{definition}{Definition}

\usepackage{chngcntr}
\usepackage{apptools}
\AtAppendix{\counterwithin{theorem}{section}}
\AtAppendix{\counterwithin{proposition}{section}}
\AtAppendix{\counterwithin{corollary}{section}}
\AtAppendix{\counterwithin{lemma}{section}}

\def\vone{{\mathbbm{1}}}

\def\gF{{\mathcal{F}}}
\def\gG{{\mathcal{G}}}

\def\gL{{\mathcal{L}}}

\def\gX{{\mathcal{X}}}
\def\gY{{\mathcal{Y}}}

\def\sR{{\mathbb{R}}}


%% file: files/1_intro.tex
\section{Introduction}

In an ever-changing world, real-time decision making necessitates coping with arbitrary distribution shifts and adversarial behavior. These shifts can arise from seasonality, change in the data distribution induced by feedback loops or policy changes, and exogenous shocks such as pandemics or economic crises. Online learning is a powerful framework for analyzing sequential data that makes no assumptions on the data distribution.   

Multi-objective learning is a generic framework that refers to any task in which a predictor must satisfy multiple objectives or criterion of interest simultaneously \citep{lee2022online}. In the online setting, this encompasses many previously studied problems such as multicalibration~\citep{pmlr-v80-hebert-johnson18a}, multivalid conformal prediction~\citep{gupta_et_al:LIPIcs.ITCS.2022.82}, and multi-group learning~\citep{deng2024groupwise}. Despite being a desirable and promising notion, methods from the online multi-objective learning literature have had little influence on the practice of machine learning. 

We attribute this to two shortcomings. First, many of the algorithms proposed in the literature are not adaptive to abrupt changes in the data distribution: they learn a predictor that minimizes the objectives over the \textit{entire time horizon}. In changing environments and in the presence of adversarial behavior, such algorithms will fail to cope with distribution shifts. Second, most prior work is purely theoretical with scant empirical evaluation. As a result, the practical aspects of multi-objective online algorithms have received limited consideration.

In this work, we aim to overcome the above shortcomings. We propose a locally adaptive multi-objective learning algorithm that outputs predictors which (approximately) satisfy a set of objectives over all local time intervals $I \subseteq [T]$. Previously, \citet{lee2022online} suggested a method that lends adaptivity to existing algorithms by including additional objectives for all contiguous subintervals. We present an alternative approach that directly modifies the multi-objective learning algorithm by replacing one part of the scheme with an adaptive online learning method. We provide a meta-algorithm that, given an adaptive online learner, minimizes the worst case multi-objective loss across time intervals. For concreteness, we instantiate it with the Fixed Share method \citep{10.1023/A:1007424614876}, which is guaranteed to provide adaptivity over all intervals of a fixed target width. Other possible instantiations of our approach that target alternative adaptive guarantees are discussed in Section \ref{sec:algorithm}. 

To close the empirical gap in this literature, we provide extensive experiments evaluating the performance of various adaptive methods in practice. This includes experiments on electricity demand forecasting and predicting recidivism over time in which our goal is to remove biases present in existing baseline predictors. Across all our empirical benchmarks we find that our proposed method consistently outperforms the previous proposals of \citet{lee2022online}. We release a codebase that implements our algorithm and all the baselines used in the paper.\footnote{\url{https://github.com/jivatneet/adaptive-multiobjective}}

As we discussed above, multi-objective learning can be used to address many common prediction tasks. As a case study, in this work, we focus on the multiaccuracy problem in which the goal is to learn predictors which are simultaneously unbiased under a set of covariate shifts of interest. We seek a small multiaccuracy error while also preserving predictive accuracy relative to a given sequence of baseline forecasts. This is a problem of significant and broad interest across real-time decision-making and deployed machine learning systems. We show that our proposed algorithm has low multiaccuracy error over contiguous subintervals while the baselines have poor adaptivity. An alternative objective to multiaccuracy that is popular in the literature is multicalibration~\citep{haghtalab2023a, doi:10.1137/1.9781611977912.98}. Despite being a stronger condition, we show that in practice existing online multicalibration algorithms only achieve multiaccuracy at relatively slow rates. Adaptive extensions of the multicalibration algorithm yield improvements in local multiaccuracy error, however are unable to close this performance gap. 

We note that although we focus on multiaccuracy in this paper, our general algorithm extends to other multi-objective learning problems including multi-group learning~\citep{pmlr-v162-tosh22a} and omniprediction~\citep{gopalan_et_al:LIPIcs.ITCS.2022.79}. We discuss these extensions in Section~\ref{sec:extensions}.

\begin{figure}[t!]
    \centering
    \begin{subfigure}[b]{0.45\textwidth}
        \includegraphics[width=\linewidth]{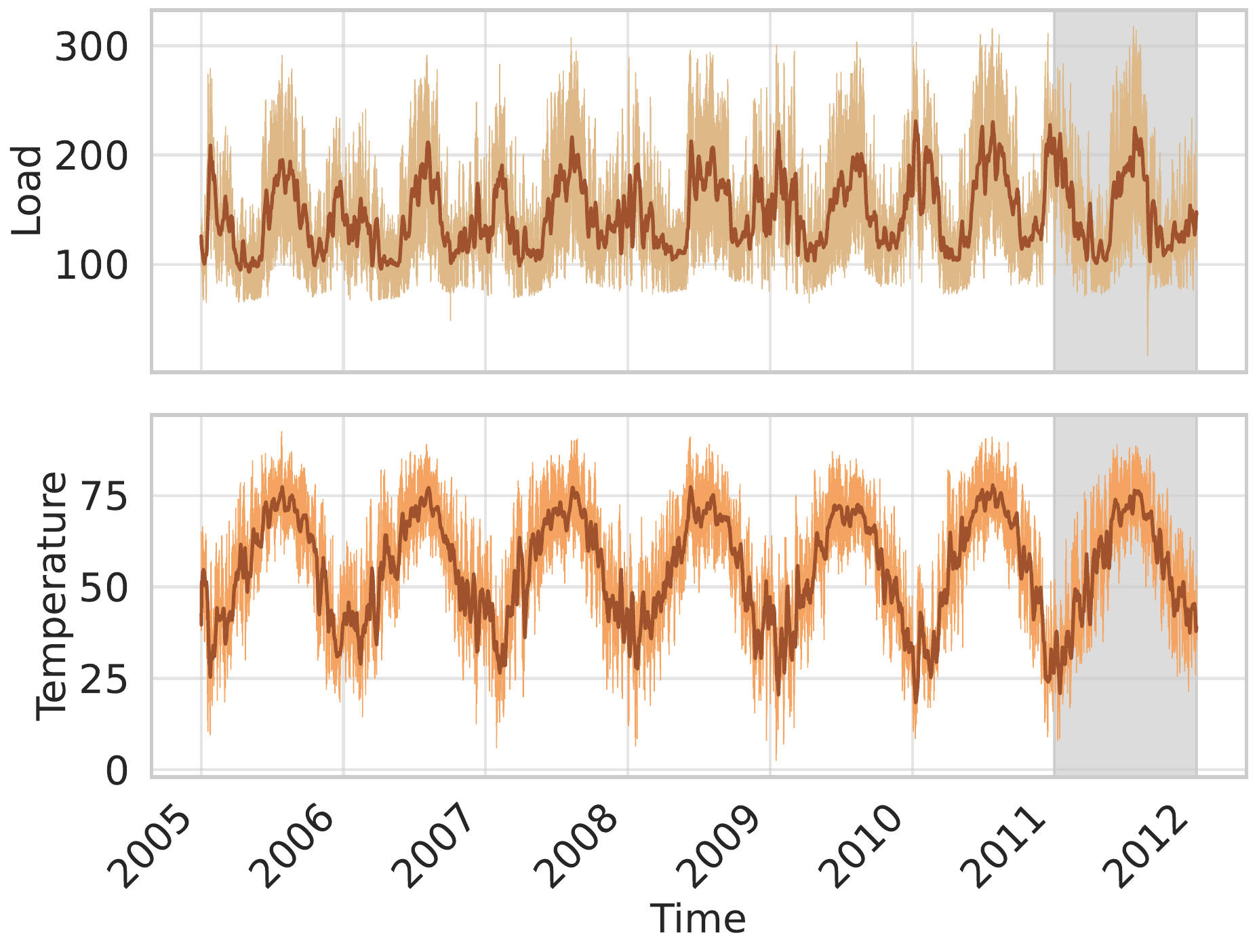}
        \caption{}
        \label{fig:load_series}
    \end{subfigure}
    \hfill
    \begin{subfigure}[b]{0.53\textwidth}
        \includegraphics[width=\linewidth]{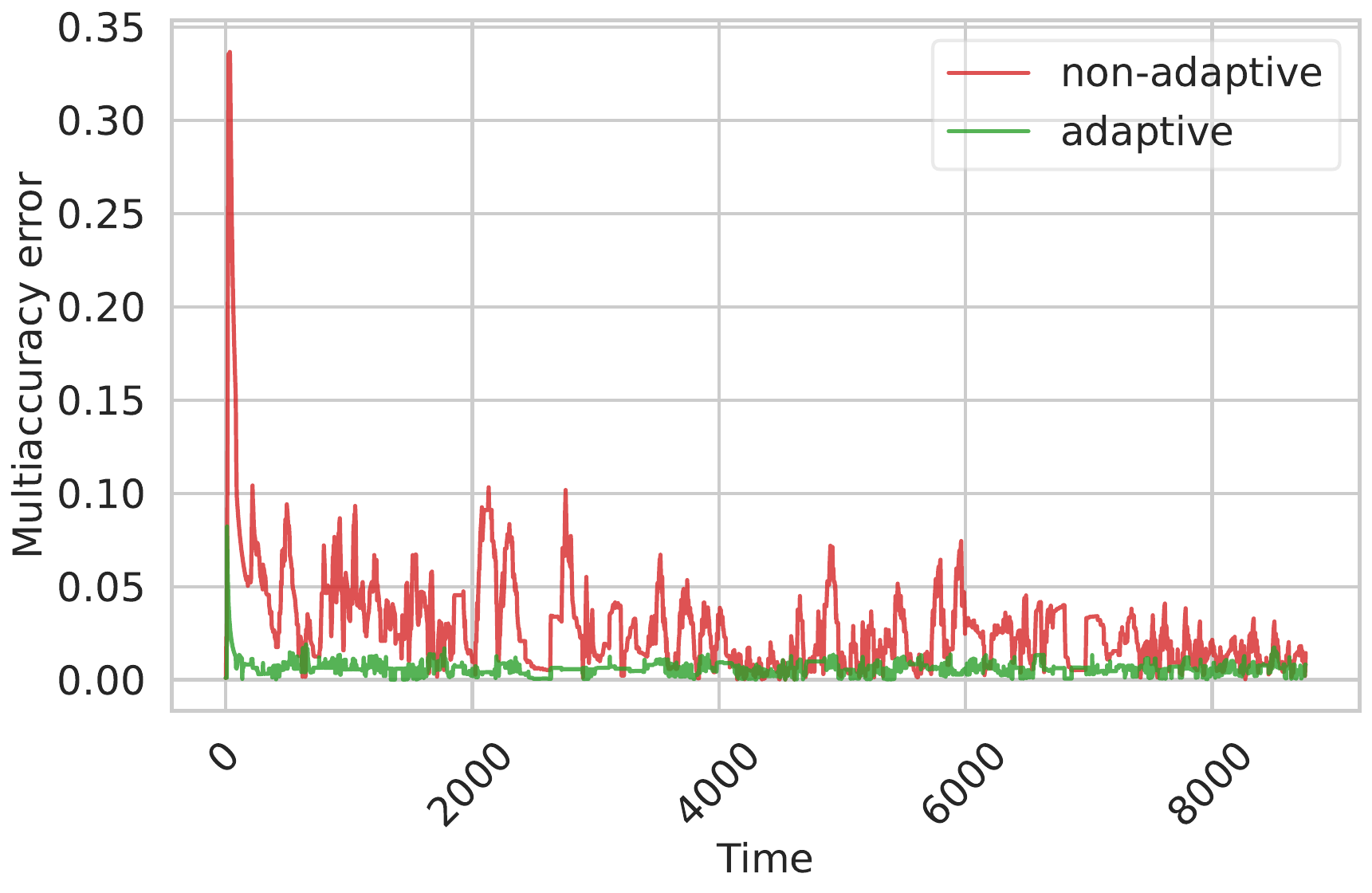}
        \caption{}
        \label{fig:peek_ma}
    \end{subfigure}
   
    \caption{\textbf{GEFCom14-L electric load forecasting dataset.} On the left hand side are the time series for the raw load (light brown) and temperature (light orange) data. The dark brown curves indicate the weekly (168-hourly) moving average. The shaded grey region shows the competition duration. On the right-hand side, we plot a weekly moving average of the local multiaccuracy error.}
    \label{fig:peek}
\end{figure}

\subsection{Peek at results}
\label{sec:peek}

To demonstrate the significance of local adaptivity in practice, we consider the probabilistic electricity load forecasting track of the Global Energy Forecasting Competition 2014 (GEFCom2014)~\citep{HONG2016896}. The aim in the load forecasting track GEFCom2014-L is to forecast month-ahead quantiles of hourly loads for a US utility from January 1, 2011 to December 31, 2011 based on historical load and temperature data (Figure~\ref{fig:load_series}).

We consider the binary task of predicting whether the electricity demand exceeds 150MW at hour $t$ and evaluate whether the predictions are multiaccurate with respect to discrete temperature groups $\{[0, 20),[20,40),\dots,[80,100)\}$ (in $^{\circ}$F). Informally, obtaining multiaccuracy with respect to temperature ensures our predictions are accurate at different times of day and across seasons. Figure~\ref{fig:peek_ma} shows the multiaccuracy error of our proposed locally adaptive algorithm compared to a non-adaptive multiaccuracy algorithm, plotted as a weekly (168-hourly) moving average. We can see that the multiaccuracy error of the adaptive algorithm is close to zero across all time intervals, while the non-adaptive variant realizes much larger errors at many time points.

\subsection{Related work}

Our work is most closely related to the literature on multi-objective learning. This encompasses numerous problems including multicalibration \citep{pmlr-v80-hebert-johnson18a}, multiaccuracy \citep{10.1145/3306618.3314287}, multi-group learning~\citep{pmlr-v162-tosh22a}, and omniprediction \citep{gopalan_et_al:LIPIcs.ITCS.2022.79}. Each of these multi-objective criteria have been studied in both the online and batch settings. Most closely related to our work, \citet{10.1145/3306618.3314287} and \citet{pmlr-v202-globus-harris23a} give algorithms for obtaining multi-accurate and multi-calibrated (respectively) predictors in the batch setting that are guaranteed to have accuracy no worse than that of a given base predictor. 

In the online adversarial setting, a number of works develop algorithms for obtaining multiaccuracy, multicalibration, and/or omniprediction globally over all time steps \citep{lee2022online, doi:10.1137/1.9781611977912.98, okoroafor2025nearoptimalalgorithmsomniprediction, haghtalab2023a, Noarav2025}. Our work will in particular build on the algorithmic framework developed in \citet{lee2022online}. This methodology has deep roots in the online learning literature and builds on ideas arising from Blackwell approachability \citep{pjm/1103044235} and its connection to no-regret learning \citep{AbernethyEtal11}.

To obtain time-local guarantees we will draw on the literature on adaptive regret~\citep{10.1023/A:1007424614876, pmlr-v37-daniely15, pmlr-v54-jun17a, haghtalab2023calibrated}. Our work will most closely rely upon the work of \citet{pmlr-v211-gradu23a} to obtain multi-objective error bounds over any local time interval. In the context of multi-objective learning, local guarantees have been discussed previously in \citet{lee2022online}. However, the literature contains no empirical evaluations of these methods. We provide experiments evaluating the algorithms of \citet{lee2022online} in Section \ref{sec:expts} and find that our approach achieves significantly lower error rates in practice.

\subsection{Preliminaries}

We use $\gX$ to denote the feature space and $\gY = [a,b]$ to denote the label space, which we assume to be a bounded interval. Our goal is to learn a sequence of predictors $p_t(x_t) \in \gY, t=1,2,\dots,T$ that achieve a low loss simultaneously for every objective within a set $\gL$ over time. Each objective, or loss, is a function $\ell: \gY \times \gX \times \gY \rightarrow [-1, 1]$ that takes as input a prediction $p$, features $x \in \gX$, and label $y \in \gY$ and returns a value in $[-1, 1]$. We will use $[T]$ to denote the set $\{1, 2,\dots,T\}$. The sequence of data points $(x_t,y_t),\, t \in [T]$ can be generated adversarially dependent on the entire history of data and predictions up to time $t$.

The objectives we consider can be quite general and we will give some examples of specific choices shortly. Broadly, our only restriction is that the objectives should be consistent with one another in the sense that for any distribution on $y_t$ there is a single optimal prediction $p_t(x_t)$ that minimizes all the objectives simultaneously. Formally, we assume the following. 

\begin{assumption}\label{assump:consistent_objs} For any $x \in \mathcal{X}$ and distribution $P_Y$ on $\mathcal{Y}$ there exists $p^* \in \gY$ such that for all $\ell \in \mathcal{L}$,
\[
p^* \in \underset{p \in \gY}{\textnormal{argmin}}\; \mathbb{E}_{Y \sim P_Y}[\ell(p,x,Y)].
\]
Moreover, for all $\ell \in \mathcal{L}$, $p^*$ guarantees the loss bound
\begin{equation}\label{eq:opt_performance}
\mathbb{E}_{Y \sim P_Y}[\ell(p^*,x,Y)] \leq 0.
\end{equation}
\end{assumption}

The assumption that $p^*$ produces a negative objective value is not strictly necessary and previous work in multiobjective learning has considered slightly more general settings \citep{lee2022online}. We have chosen to add this condition because it simplifies the notation and is satisfied by many common problems of interest. For instance, as we will discuss in the sections that follow, multiaccuracy, multicalibration, omniprediction, and multi-group learning can all be formulated in a way that meets this condition.  

Using this assumption, our goal in online multi-objective learning will be to learn a sequence of predictions $p_t(x_t)$ that (approximately) matches the optimal bound (\ref{eq:opt_performance}):
\[
\underset{\ell \in \gL}{\operatorname{max}} \;\frac{1}{T} \sum_{t=1}^T  \ell(p_t(x_t), x_t, y_t) \lessapprox 0.
\]

As examples, we now define two instantiations of multi-objective problems that are commonly studied in the literature and which we will focus on—multiaccuracy and multicalibration. The offline version of multiaccuracy was introduced in~\citet{10.1145/3306618.3314287}. We parameterize the multiaccuracy criterion by a function class $\gF$ and the goal is to be unbiased for all $f \in \gF$, i.e., there is no systematic correlation between the prediction residuals and any $f \in \gF$.

\begin{definition}[Online multiaccuracy] Let $\gF = \{f: \gX \rightarrow [0,1]  \}$ be a class of functions on $\gX$. In online multiaccuracy, we instantiate $\ell_{\ma_{f,\sigma}}(p_t(x_t),x_t, y_t) = \sigma f(x_t) \cdot (y_t - p_t(x_t))$ for every sign $\sigma = \{\pm\}$ and $f \in \gF$ and define the multiaccuracy error $\ell_\ma$ in the sup-norm as 

\begin{equation}
\label{eq:ma_error}
    \ell_\ma(p_t(x_t),x_t, y_t) = \underset{f \in \gF, \sigma \in \{\pm\}}{\operatorname{sup}} \;\frac{1}{T} \sum_{t=1}^T \sigma f(x_t) \cdot (y_t - p_t(x_t)).
\end{equation}
    
\end{definition}

Another popular online prediction target is multicalibration \citep{pmlr-v80-hebert-johnson18a}. In a binary classification task, calibration asks that among instances with predicted probability $p$, a fraction $p$ of them are observed to be truly labeled as 1. Multicalibration is a strengthening of calibration that additionally requires the predictor to be multiaccurate conditional on its realized value. To implement this in practice, we discretize the label interval $[0, 1]$ into $m$ bins $V_m := \{[0, 1/m),[1/m, 2/m),\dots, [(m-1)/m, 1] \}$ and define a representative value for each bin as the midpoint
$v_j=\tfrac{2j-1}{2m}$ for $j=1,\dots,m$. We then define an approximate notion of multicalibration that asks for $v_j$ to be an unbiased prediction of $y_t=1$ over all reweightings in $\mathcal{F}$ and all timepoints where $p_t(x_t) \in [v_j - \frac{1}{2m}, v_j + \frac{1}{2m})$.  

\begin{definition}[Online multicalibration] Fix a set of functions $\gF$ and $m \geq 1$. In online multicalibration we instantiate $\ell_{\mathrm{MC}_{f,\sigma,v}}(p_t(x_t),x_t, y_t) = \sigma f(x_t) \cdot \vone\{p_t(x_t) \in v\} \cdot (y_t - v_j)$ for every sign $\sigma = \{\pm\}$, $f \in  \gF$, and $v \in V_m$ and define the multicalibration error $\ell_\mc$ in the sup-norm as 
\begin{equation}
    \ell_\mc(p_t(x_t),x_t, y_t) = \underset{f \in \gF, \sigma \in \{\pm\}, v \in V_m}{\operatorname{sup}} \;\frac{1}{T} \sum_{t=1}^T \sigma f(x_t) \cdot \vone\{p_t(x_t) \in v\} \cdot (y_t - v_j).
\end{equation}
\end{definition}

A direct calculation shows that the online multicalibration error always upper bounds the multiaccuracy error; specifically, $\ell_\ma \leq m \cdot \ell_\mc + 1/(2m)$. 

In this work, we will give a multi-objective learning algorithm that achieves small multiaccuracy error while preserving predictive accuracy relative to a base predictor sequence $\tilde{p}_t(x_t), t \in [T]$. While improving multiaccuracy, it is important that we do not degrade the predictive accuracy of $\tilde{p}_t(x_t)$, leaving its forecasts less useful. We discuss this in more detail in Section~\ref{sec:acc_objective_significance}. We define the latter accuracy objective as prediction error. In what follows, we let $c : \mathcal{Y} \times \mathcal{Y} \to \mathbb{R}_{\geq 0}$ denote any proper loss for the mean, i.e., any loss such that $\mathbb{E}_{y \sim P}[y] \in \text{argmin}_p \mathbb{E}_{y \sim P}[c(p,y)]$ for all distributions $P$ on $\mathcal{Y}$. A common example that we will work with in our experiments is the squared error/Brier score $c(p,y) = (y-p)^2$. 
 
\newpage
\begin{definition}[Online prediction error] Given a base predictor sequence $\tilde{p}_t(x_t), t \in [T]$, define the prediction error $\ell_\pred$ as

\begin{equation}
\label{eq:lpred}
    \ell_\pred(p_t(x_t),x_t, y_t):= \frac{1}{T} \sum_{t=1}^T c(p_t(x_t), y_t) - c(\tilde{p}_t(x_t), y_t),
\end{equation}
\label{def:pred}
\end{definition}

%% file: files/2_method.tex
\section{Methods}

\subsection{Online multi-objective learning}\label{sec:mo}

The online multi-objective learning problem is a sequential prediction task over $T$ rounds. A standard framework introduced in \citet{lee2022online} is to consider a two-player game between a learner, who observes $x_t \in \gX$ and chooses a predictor $p_t(x_t)$, and an adversary who maintains a distribution $q^{(t)} \in \Delta(\gL)$, where we use the notation $\Delta(S)$ to denote the set of probability distributions over the set $S$. At each time step, the learner observes the adversary's current mixture and the covariates $x_t$ and chooses its (randomized) prediction as
\[
p_t(x_t) \sim P_t(x_t), \ \ \  \text{ where } \ \ \  P_t(x_t) = \underset{P \in \Delta(\gY)}{\operatorname{argmin}}\; \underset{y \in \gY}{\operatorname{max}}\; \mathbb{E}_{p \sim P}\left[ \sum_\ell q_{\ell}^{(t) } \ell(p, x_t, y) \right].
\]
This choice is designed to guarantee that the learner obtains the best possible performance under the adversarial value of $y_t$ with respect to the mixture loss specified by $q^{(t)}$. As an aside, we note that although generic multiobjective learning problems require randomized predictors, our methods will often produce deterministic values. This is due to the fact that for many of the problems we are interested in (e.g., multiaccuracy, low predictive accuracy) the objectives are convex and thus the minimax program above admits a  solution $P_t(x_t)$ that is supported on a singleton. 

After the learner makes its selection, the true value of $y_t$ is revealed and the adversary updates its mixture distribution. In the original work of \cite{lee2022online}, the adversary sets its weights using the Hedge updates
\[
q^{(t+1)}_{\ell} \propto q^{(t)}_{\ell} \exp(\eta \ell(p_t(x_t),x_t,y_t)),
\]
for some $\eta = \Theta(\sqrt{\log(|\mathcal{L}|)/T})$. This is designed to ensure that the mixture distribution with respect to $q^{(t)}$ is a good proxy for the maximum multiobjective error. More formally, this choice of weights has the following well-known error bound (see, e.g., Theorem 1.5 of \citet{hazan2016introduction}),
\[
\max_{\ell \in \mathcal{L}} \sum_{t=1}^T \mathbb{E}_{p \sim P_t(x_t)} [\ell(p,x_t,y_t)] \leq \sum_{\ell \in \mathcal{L}} \sum_{t=1}^T q_{\ell}^{(t)} \mathbb{E}_{p \sim P_t(x_t)} [\ell(p,x_t,y_t)] + O(\sqrt{T\log(|\mathcal{L}|)}).
\]
By combining this bound with the above choice of $p_t(x_t)$ we obtain the following multiobjective error bound.
\begin{theorem}[Theorem 2.1 in \citet{lee2022online}]\label{thm:lee}
    Under Assumption \ref{assump:consistent_objs}, Algorithm \ref{alg:generic_alg} with Hedge as the method for learning $q^{(t)}$ obtains the multiobjective learning bound
    \[
    \max_{\ell \in \mathcal{L}} \sum_{t=1}^T \mathbb{E}_{p \sim P_t(x_t)} [\ell(p,x_t,y_t)] \leq O(\sqrt{T\log(|\mathcal{L}|)}).
    \]
\end{theorem}

\subsection{Locally adaptive multi-objective learning}
\label{sec:algorithm}

The result of Theorem \ref{thm:lee} ceases to be useful when environments are changing and the data distribution shifts over time. As a simple example, fix the singleton function class $\mathcal{F}_{\mathrm{MA}} = \{x\mapsto 1\}$ and consider targeting just the multiaccuracy error (i.e., set $\gL = \{\ell_{\ma_{f,\sigma}} : f \in \mathcal{F}_{\mathrm{MA}}, \sigma \in \{\pm \}\}$). Let the labels be given as $y_t=1$ for the first $T/2$ rounds and $y_t = 0$ for the last $T/2$ rounds. Here, the constant predictor $p_t = 1/2$ minimizes the multiaccuracy error in (\ref{eq:ma_error}). Nevertheless, this predictor performs poorly in the individual intervals $1 \leq t \leq T/2$ and $t > T/2$ compared to the optimal predictor that switches from $p_t = 1$ to $p_t = 0$ after $t = T/2.$

To account for distribution shifts in changing environments, we will now modify the method of \citet{lee2022online} by replacing the Hedge algorithm with a locally adaptive method. Informally, this will allow us to bound the worst case multi-objective loss over local subintervals given by
\begin{align}
\label{eq:local_regret}
     \underset{I = [r,s] }{\operatorname{sup}} \Bigg[ \underset{\ell \in \gL}{\operatorname{max}} \sum_{t=r}^s \mathbb{E}_{p \sim P_t(x_t)} [\ell(p, x_t, y_t)]\Bigg],
\end{align}
where the supremum is over some appropriate set of intervals $I$ that we will specify shorty. Algorithm \ref{alg:generic_alg} gives our generic method. Here, \texttt{WL} denotes any procedure for learning the weights $q^{(t)}$. 

\begin{algorithm}[ht]
\caption{Locally adaptive multi-objective learning}
\label{alg:generic_alg}
\begin{algorithmic}[1]
\REQUIRE Set of objectives $\gL$, learning method \texttt{WL} 
\REQUIRE Sequence of samples $\{ (x_1, y_1), \ldots, (x_T, y_T) \}$ 
\STATE $q_{\ell}^{(1)} =\frac{1}{|\gL|}, \quad \forall \ell \in \mathcal{L}$.
\FOR{each $t \in [T]$}
    \STATE $P_t(x_t) =  \underset{P \in \Delta(\gY)}{\operatorname{argmin}} \;\underset{y \in \gY}{\operatorname{max}} \;\mathbb{E}_{p \sim P}\left[  \sum\limits_{\ell \in \gL} q_{\ell}^{(t)} \ell(p, x_t, y_t) \right]$
    \STATE Output $p_t(x_t) \sim P_t(x_t)$
    \STATE $q_{\ell}^{(t+1)} = \texttt{WL}(\{q^{(s)}\}_{s \leq t}, \{\mathbb{E}_{p \sim P_t(x_t)}[\ell(p,x_t,y_t)]\}_{\ell \in \mathcal{L}})$
    \label{line:finite-ma-exp}
\ENDFOR
\end{algorithmic}
\end{algorithm}

As a concrete instantiation, we will perform empirical experiments on the Fixed Share method introduced in \cite{10.1023/A:1007424614876} that modifies the Hedge update by adding an exploration term that prevents any of the  weights from collapsing to zero. A formal statement of this procedure is given in Algorithm \ref{alg:fixed-share}. As we will discuss in the next section, Fixed Share provides a multiobjective learning guarantee \textit{locally} on any interval of a fixed width. There are many possible alternative methods that one could implement in the place of Fixed Share. For instance, one may consider the \textit{strongly adaptive} learning procedure of \citet{pmlr-v37-daniely15} and \citet{pmlr-v54-jun17a} that guarantee a stronger notion of adaptive regret with dependency over the interval width $|I|$ for all intervals $I \subseteq [T]$. We have chosen to focus on Fixed Share due to its strong empirical performance.

\begin{algorithm}[H]
\caption{Fixed-Share weight update}
\label{alg:fixed-share}
\begin{algorithmic}[1]
\REQUIRE Weights at current timestep $q^{(t)}$; hyperparameters $\eta, \gamma$. 
\REQUIRE Losses for current timestep $\{ \mathbb{E}_{p \sim P_t(x_t)}[\ell(p,x_t,y_t)] \}_{\ell \in \mathcal{L}}$ 
\FOR{each $t \in [T]$}
    \STATE $\tilde{q}_{\ell}^{(t+1)} = \dfrac{ q_{\ell}^{(t)} \exp\left( \eta \cdot \mathbb{E}_{p \sim P_t(x_t)}[\ell(p,x_t,y_t)] \right)}{\sum_{\ell' \in \mathcal{L}} q_{\ell'}^{(t)} \exp\left( \eta \cdot \mathbb{E}_{p \sim P_t(x_t)}[\ell'(p,x_t,y_t)] \right)}$, for all $\ell \in \mathcal{L}$
    \STATE $q_{\ell}^{(t+1)} = (1 - \gamma) \, \tilde{q}_{\ell}^{(t+1)} + \frac{\gamma}{|\mathcal{L}|}$
    \label{line:finite-ma-exp}
\ENDFOR
\ENSURE Weights for the next time step $q^{(t+1)}$
\end{algorithmic}
\end{algorithm}

\paragraph{Comparison to adaptive algorithms in literature.} Previously,~\citet{lee2022online} proposed an adaptive extension of their multi-objective learning algorithm that included additional objectives for all subintervals. Formally, given an initial set of objectives $\mathcal{L}$ they consider the augmented collection $\mathcal{L}_{\text{adapt.}} = \{\ell(p_t(x_t),x_t,y_t) \mathbbm{1}\{t \in I\} \mid  \ell \in \mathcal{L}, I = [r,s] \subseteq [T]\}$ and show that using these objectives in the algorithm described in Section \ref{sec:mo} guarantees the local bound
\[
\sup_{I = [r,s] \subseteq [T]} \Bigg[ \underset{\ell \in \gL}{\operatorname{max}} \sum_{t \in [I]} \mathbb{E}_{p \sim P_t(x_t)} [\ell(p,x_t,y_t)] \Bigg] \leq O(\sqrt{T(\log(|\mathcal{L}|) + \log T)},
\]
where the supremum is over all contiguous intervals $I \subseteq [T]$. In our work, we propose to instead hold the set of objectives fixed and  use a locally adaptive procedure \texttt{WL} to learn the weights $q^{(t)}$. 

\section{Theory}
\label{sec:theory}

We will now state a theoretical guarantee for Algorithm \ref{alg:generic_alg}. For concreteness, we will focus on the case where the adversary learns the weights $q^{(t)}$ using the Fixed Share method given in Algorithm \ref{alg:fixed-share}. Similar results for other adaptive learning methods can be obtained in an identical fashion by replacing the regret bound for Fixed Share (Lemma \ref{lem:expert_interval_regret} below) with the associated bound for that method. 

The theory has two parts: a guarantee for the adversary's distribution $q$ and a guarantee on the learner's response. From here on, we use the shorthand $\ell^{(t)} := \mathbb{E}_{p \sim P_t(x_t)}[\ell(p, x_t, y_t)]$ to denote the expected loss of our randomized predictor at time step $t$ and denote the $|\gL|$-dimensional vector of losses as $\ell^{(t)}_\gL = (\ell^{(t)})_{\ell \in \gL}$. All proofs are deferred to Appendix~\ref{sec:proofs}.

We first show that the maximum objective value over any time interval $I$ is upper bounded by the average value of the individual objectives taken with respect to the weights $q^{(t)}$.

\begin{lemma} 
\label{lem:expert_interval_regret}
Consider Algorithm \ref{alg:generic_alg} with weights learned using Algorithm \ref{alg:fixed-share}. Assume that $\gamma \leq 1/2$ and $\eta \leq 1$. Then, for any interval $I = [r, s] \subseteq [T]$, 
\begin{equation}
\label{eq:lem1_mo}
    \hspace{-3mm}
     \sum_{t=r}^s  q^{(t)\top} \ell^{(t)}_\gL \geq  \underset{\ell \in \gL}{\operatorname{max}} \sum_{t=r}^s \ell^{(t)} - \eta \Bigg(\sum_{t=r}^s q^{(t)\top} ({\ell^{(t)}_\gL})^2\Bigg) - \frac{1}{\eta}\bigg(\log\bigg(\frac{|\mathcal{L}|}{\gamma} \bigg) + |I|2\gamma \bigg),
\end{equation}
where we use the notation $(\ell^{(t)}_{\mathcal{L}})^2$ to denote the elementwise square of the vector $\ell^{(t)}_{\mathcal{L}}$.

\end{lemma}

Next, we show that the average value of the objectives is non-positive over any interval $I$. This lemma follows from the minimax-optimal strategy of the learner and has been shown to hold previously in~\citet{lee2022online}.

\begin{lemma} Suppose the objectives satisfy Assumption \ref{assump:consistent_objs}. Then, for any interval $I = [r, s] \subseteq [T]$,
\label{lem:expected_l_le_0}
\[
     \sum_{t=r}^s  q^{(t)\top} \ell^{(t)}_\gL \leq 0.
\]
\end{lemma}

We combine the previous two lemmas to get our main result.

\begin{theorem}
\label{thm:adaptive_loss_bound}
Fix any $\gamma \leq 1/2$ and $\eta \leq 1$ and assume that the objectives satisfy Assumption \ref{assump:consistent_objs}. Then, for any interval $I = [r, s] \subseteq [T]$, 
\begin{equation}
\label{eq:thm1_mo_bound}
    \underset{\ell \in \gL}{\operatorname{max}} \frac{1}{|I|} \sum_{t=r}^s \ell^{(t)} \leq \frac{\eta}{|I|} \Bigg(\sum_{t=r}^s q^{(t)\top} ({\ell^{(t)}_\gL})^2\Bigg) + \frac{1}{\eta |I|}\bigg(\log\bigg(\frac{|\mathcal{L}|}{\gamma} \bigg) + |I|2\gamma \bigg).
\end{equation}

\end{theorem}

The guarantee of Theorem~\ref{thm:adaptive_loss_bound} depends on the values of the fixed share hyperparameters $\gamma, \eta$. To set the best upper bound for a given interval $I$, we would ideally substitute the optimal values $\gamma = \dfrac{1}{2|I|}$ and $\eta = \sqrt{\dfrac{\log(|\mathcal{L}|\cdot 2|I|) + 1}{\sum_{t=r}^s q^{(t)\top} ({\ell^{(t)}_\gL})^2}}$ in (\ref{eq:thm1_mo_bound}) and obtain
\begin{equation}
\label{eq:mo_final_upper_bound}
   \underset{\ell \in \gL}{\operatorname{max}} \frac{1}{|I|} \sum_{t=r}^s \ell^{(t)} \leq \frac{2}{|I|} \sqrt{\Bigg(\log(|\mathcal{L}|\cdot 2|I|) + 1 \Bigg)} \cdot \sqrt{\sum_{t=r}^s q^{(t)\top} ({\ell^{(t)}_\gL})^2}= O \Bigg(\sqrt{\frac{\log(|\gL|\cdot |I|)}{|I|}}\Bigg).
\end{equation}

In practice, we can only use one setting of these parameters and cannot specialize $\gamma$ and $\eta$ to a specific interval. To mimic these optimal choices, we let the user pick a fixed target interval width $|I| = \tau$, noting that a smaller choice of $\tau$ gives stronger locally adaptive guarantees at the cost of a looser upper bound. Since the optimal value for $\eta$ used above depends on the expected squared loss $\sum_{t=r}^s q^{(t)\top} ({\ell^{(t)}_\gL})^2$ which is unknown in practice, we follow~\citet{JMLR:v25:22-1218} in selecting an adaptive value of $\eta$ that updates online as
\begin{equation}
\label{eq:adaptive_eta}
\eta = \eta_t := \sqrt{\frac{\log(|\mathcal{L}|\cdot 2\tau) + 1}{\sum_{s=t-\tau +1}^{t} q^{(s)\top} ({\ell^{(s)}_\gL})^2}}.
\end{equation}

This choice lets the algorithm adaptively track changes in the moving average of the expected squared loss over the most recent $\tau$ time steps.

\section{Applications to Mean Estimation and Quantile Estimation}

We now consider two example applications of Algorithm \ref{alg:generic_alg} to multiaccurate mean and quantile estimation.

\subsection{Multiaccurate mean estimation}
\label{sec:multiaccuracy_case}

As a case study, we focus on the multiaccuracy problem in this work. Our goal is to learn predictors that have small multiaccuracy error (\ref{eq:ma_error}) while guaranteeing the prediction error (\ref{eq:lpred}) is low relative to a given sequence of baseline predictions $\{\tilde{p}_t(x_t)\}_{t=1}^T$. 
We fix a function class $\gF_{\ma} \subseteq \{f: \gX \rightarrow [0,1] \}$ that we desire multiaccuracy with respect to and define $\gL := \{\ell_{\ma_{f,\sigma}} : f \in \mathcal{F}_{\text{MA}}, \sigma \in \{\pm \} \} \cup\ \{\ell_{\pred}\}$. We use the shorthands $\ell_{\ma_{f,\sigma}}^{(t)} := \sigma f(x_t)(y_t - p_t(x_t))$ and $\ell_{\pred}^{(t)} := c(p_t(x_t), y_t) - c(\tilde{p}_t(x_t), y_t)$ to denote the realized losses. Assuming $c$ is convex, we note that since the objectives $\ell_{\ma_{f,\sigma}}$ and $\ell_{\pred}$ are convex we may assume without loss of generality that the prediction $p_t(x_t)$ is deterministic. We provide an algorithm for locally adaptive multiaccurate mean estimation in Algorithm~\ref{alg:multiobjective-ma-regret} and its guarantee in Corollary~\ref{cor:multiaccuracy}. The weights $q^{(t)}_{\text{MA},f,\sigma}$ and $q_{\pred}^{(t)}$ in Algorithm~\ref{alg:multiobjective-ma-regret} are used to denote the entries of $q^{(t)}$ associated with the multiaccuracy and prediction error objectives, respectively. 

\begin{corollary} 
\label{cor:multiaccuracy}
Consider the weights learned using Algorithm \ref{alg:multiobjective-ma-regret} with $\eta = \Theta\left(\sqrt{\frac{\log((|\mathcal{F}_{\textup{MA}}| + 1)\cdot \tau)}{\tau}}\right) \leq 1$ and $\gamma = 1/(2\tau)$ for some $\tau \geq 1$. Then, for any interval $I = [r, s] \subseteq [T]$ of length $|I| = \tau$, 
\[
 \max \Bigg\{ \underset{f, \sigma}{\operatorname{max}}  \frac{1}{|I|}\sum_{t=r}^{s} \ell_{\nma_{f,\sigma}}^{(t)}, \frac{1}{|I|}\sum_{t=r}^{s} \ell_{\npredt}^{(t)}\Bigg\} \leq O \Bigg(\sqrt{\frac{\log((|\mathcal{F}_{\textup{MA}}| + 1)\cdot \tau)}{\tau}}\Bigg).
\]
\end{corollary}

\begin{algorithm}[ht]
\caption{Locally adaptive multiaccurate mean estimation}
\label{alg:multiobjective-ma-regret}
\begin{algorithmic}[1]
\REQUIRE Function class $\gF_{\ma} \subseteq \{f: \gX \rightarrow [0,1] \}$; base predictor sequence $\tilde{p}_t(x_t), t \in [T]$; \\hyperparameters $\eta, \gamma$. 
\REQUIRE Sequence of samples $\{ (x_1, y_1), \ldots, (x_T, y_T) \}$ 
\STATE $q_{{\ma}_{f, \sigma}}^{(1)} =\frac{1}{2|\mathcal{F}_\ma| + 1}, \quad \forall f \in \mathcal{F}_{\ma}, \sigma \in \{\pm 1\}$.
\STATE $q_{\pred}^{(1)} =\frac{1}{2|\mathcal{F}_\ma| + 1}$
\FOR{each $t \in [T]$}
    \STATE $p_t(x_t) :=  \underset{p \in \mathcal{Y}}{\operatorname{argmin}} \;\underset{y \in \gY}{\operatorname{max}} \sum\limits_{f,\sigma} q_{\ma_{f,\sigma}}^{(t)} \sigma f(x_t)(y - p) + q_{\pred}^{(t)} (c(p, y) - c(\tilde{p}_t(x_t), y))$
    \STATE $\tilde{q}_{\ma_{f,\sigma}}^{(t+1)} = \dfrac{q_{\ma_{f,\sigma}}^{(t)} \exp\left( \eta \cdot \sigma f(x_t)(y_t - p_t(x_t)) \right)}{\sum_{\ell' \in \mathcal{L}} q_{\ell'}^{(t)} \exp\left( \eta \cdot \ell'(p_t(x_t),x_t,y_t) \right)}$, for all $f \in \mathcal{F}_\ma, \sigma \in \{\pm 1\}$
    \STATE $\tilde{q}_{\pred}^{(t+1)} = \dfrac{q_{\pred}^{(t)} \exp\left( \eta \cdot (c(p_t(x_t), y_t) - c(\tilde{p}_t(x_t), y_t)  )\right)}{\sum_{\ell' \in \mathcal{L}} q_{\ell'}^{(t)} \exp\left( \eta \cdot \ell'(p_t(x_t),x_t,y_t) \right)}$
    \STATE $q_{\ma_{f,\sigma}}^{(t+1)} = (1 - \gamma) \, \tilde{q}_{\ma_{f,\sigma}}^{(t+1)} + \frac{\gamma}{2|\mathcal{F}_\ma| + 1}$
    \STATE $q_{\pred}^{(t+1)} = (1 - \gamma) \, \tilde{q}_{\pred}^{(t+1)} + \frac{\gamma}{2|\mathcal{F}_\ma| + 1}$
    \label{line:finite-ma-exp}
\ENDFOR
\ENSURE Sequence of predictions $p_1(x_1), \ldots, p_T(x_T)$
\end{algorithmic}
\end{algorithm}

Next, we discuss the importance of including the prediction error objective in multiaccuracy problems.

\subsection{Significance of the prediction error objective}
\label{sec:acc_objective_significance}

In our applications, we will start with a base forecaster, $\tilde{p}_t(x_t)$ that was constructed in advance for that application. Our goal will be to improve $\tilde{p}_t(x_t)$ to be multiaccurate. While doing this, it is important that we do not degrade the accuracy of $\tilde{p}_t(x_t)$, thereby rendering its predictions less useful. Our algorithm achieves small multiaccuracy error while preserving the predictive accuracy relative to a base predictor by including an additional prediction error objective (\ref{def:pred}).

If such a base forecaster is not available, one might consider omitting the prediction error objective and running our method with just the multiaccuracy objectives $\mathcal{L} =  \{\ell_{\ma_{f,\sigma}} : f \in \mathcal{F}_{\text{MA}}, \sigma \in \{\pm \} \}$. In general, this is not advisable. Indeed, if we exclude the predictive accuracy objective in Algorithm \ref{alg:multiobjective-ma-regret} one can show that the minmax program yields the prediction: $p_t(x_t) = b\vone\big\{\sum_{f, \sigma} q_{\ma_{f,\sigma}}^{(t)}  \sigma f(x_t) > 0\big\} + a\vone\big\{\sum_{f, \sigma} q_{\ma_{f,\sigma}}^{(t)}  \sigma f(x_t) \leq 0\big\} $. This solution has the pathological behavior of only producing predictions at the extreme values $a$ or $b$ at every step. This makes the predictions less useful and interpretable for real-time decision-making in an online setting. Our prediction error objective recovers the predictor from this problem by enforcing solutions that do not lie in the extremes. In practical settings where $\tilde{p}_t(x_t)$ is not available in advance, we recommend combining our procedure with a standard online learning algorithm (e.g., online gradient or mirror descent) that provides an appropriate baseline (see, e.g., Algorithm~\ref{alg:multiobjective-ma-regret-nobaseline} in the appendix).

\subsection{Multiaccurate quantile estimation}

Our algorithm can also be employed for quantile estimation. For a user-specified quantile level $\alpha \in (0, 1)$, we seek to obtain quantile predictions $\theta_t(x_t)$ that minimize
\[
\left| \frac{1}{T}\sum_{t=1}^T \vone \{y_t \leq \theta_t(x_t)\} - \alpha \right|.
\]
We refer to this objective as \textit{coverage} and its interpretation is that $\theta_t(x_t)$ lies above $y_t$ with frequency $\alpha$. It is well known that in the population limit minimizing the quantile loss $\ell_\alpha$ (also referred to as pinball loss) produces the desired quantile predictors. Given a sequence of baseline quantile predictions $\tilde{\theta}_t(x_t)$, our goal is to update the predictions to satisfy a multiaccurate coverage criterion specified by $\gF_{\ma}$ while preserving the quantile loss $\ell_\alpha$ relative to $\tilde{\theta}_t(x_t)$. In particular, we define $\gL := \{\sigma f(x_t)(\vone \{y_t \leq \theta_t(x_t)\} - \alpha)  : f \in \mathcal{F}_{\text{MA}}, \sigma \in \{\pm \} \} \cup\ \{\ell_\alpha(\theta_t(x_t), y_t) - \ell_\alpha(\tilde{\theta}_t(x_t), y_t)\}$. We provide the explicit algorithm in Algorithm~\ref{alg:multiobjective-quantile} and its guarantee in Corollary \ref{cor:multiaccuracy_quantiles}. Note that we have to allow $\theta_t(x_t)$ to be random in this algorithm. The weights $q^{(t)}_{\text{MA},f,\sigma}$ and $q_{\pred}^{(t)}$ in Algorithm~\ref{alg:multiobjective-quantile} are used to denote the entries of $q^{(t)}$ associated with the multiaccuracy and quantile loss objectives, respectively.

\begin{corollary} 
\label{cor:multiaccuracy_quantiles}
Consider the weights learned using Algorithm \ref{alg:multiobjective-quantile} with $\eta = \Theta\left(\sqrt{\frac{\log((|\mathcal{F}_{\textup{MA}}| + 1)\cdot \tau)}{\tau}}\right) \leq 1$ and $\gamma = 1/(2\tau)$ for some $\tau \geq 1$. Then, for any interval $I = [r, s] \subseteq [T]$ of length $|I| = \tau$, 
\begin{align*}
\max \Bigg\{ \underset{f, \sigma}{\operatorname{max}}  &\frac{1}{|I|}\sum_{t=r}^{s} \mathbb{E}_{\theta \sim \Theta_t(x_t)} [\sigma f(x_t)(\vone \{y_t \leq \theta\} - \alpha)],\\ & \frac{1}{|I|}\sum_{t=r}^{s} \mathbb{E}_{\theta \sim \Theta_t(x_t)}[\ell_\alpha(\theta, y_t) - \ell_\alpha(\tilde{\theta}_t(x_t), y_t)]\Bigg\} \leq O \Bigg(\sqrt{\frac{\log((|\mathcal{F}_{\textup{MA}}| + 1)\cdot \tau)}{\tau}}\Bigg).    
\end{align*}

\end{corollary}

\begin{algorithm}[H]
\caption{Locally adaptive multiaccurate quantile estimation}
\label{alg:multiobjective-quantile}
\begin{algorithmic}[1]
\REQUIRE Function class $\gF_{\ma} \subseteq \{f: \gX \rightarrow [0,1] \}$; quantile level $\alpha$; baseline quantile predictions $\tilde{\theta}_t(x_t), t \in [T]$; hyperparameters $\eta, \gamma$. 
\REQUIRE Sequence of samples $\{ (x_1, y_1), \ldots, (x_T, y_T) \}$ 
\STATE $q_{{\ma}_{f, \sigma}}^{(1)} =\frac{1}{2|\mathcal{F}_\ma| + 1}, \quad \forall f \in \mathcal{F}_{\ma}, \sigma \in \{\pm1\}$.
\STATE $q_{\pred}^{(1)} =\frac{1}{2|\mathcal{F}_\ma| + 1}$
\FOR{each $t \in [T]$}
    \STATE $\Theta_t(x_t) :=  \underset{\Theta \in \Delta(\gY)}{\operatorname{argmin}}\; \underset{y \in \gY}{\operatorname{max}}\; \mathbb{E}_{\theta \sim \Theta} \left[\sum_{f,\sigma} q_{\ma_{f,\sigma}}^{(t)} \sigma f(x_t)(\vone \{y \leq \theta\} - \alpha) 
+ q_{\pred}^{(t)}  \left(\ell_\alpha(\theta, y) - \ell_\alpha(\tilde{\theta}_t, y_t) \right)\right]$
    \STATE Output $\theta_t(x_t) \sim \Theta_t(x_t)$
    \STATE $\tilde{q}_{\ma_{f,\sigma}}^{(t+1)} = \dfrac{q_{\ma_{f,\sigma}}^{(t)} \exp\left( \eta \cdot \mathbb{E}_{\theta \sim \Theta_t(x_t)} [\sigma f(x_t)(\vone \{y_t \leq \theta\} - \alpha)]  \right)}{\sum_{\ell' \in \mathcal{L}} q_{\ell'}^{(t)} \exp\left( \eta \cdot \mathbb{E}_{\theta \sim \Theta_t(x_t)} [\ell'(\theta,x_t,y_t)] \right)}$ for all $f \in \mathcal{F}_\ma, \sigma \in \{\pm1\}$
    \STATE $\tilde{q}_{\pred}^{(t+1)} = \dfrac{q_{\pred}^{(t)} \exp\left( \eta \cdot \mathbb{E}_{\theta \sim \Theta_t(x_t)}[(\ell_\alpha(\theta, y_t) - \ell_\alpha(\tilde{\theta_t}(x_t), y_t)]   \right)}{\sum_{\ell' \in \mathcal{L}} q_{\ell'}^{(t)} \exp\left( \eta \cdot \mathbb{E}_{\theta \sim \Theta_t(x_t)}[\ell'(\theta,x_t,y_t)] \right)}$
    \STATE $q_{\ma_{f,\sigma}}^{(t+1)} = (1 - \gamma) \, \tilde{q}_{\ma_{f,\sigma}}^{(t+1)} + \frac{\gamma}{2|\mathcal{F}_\ma| + 1}$
    \STATE $q_{\pred}^{(t+1)} = (1 - \gamma) \, \tilde{q}_{\pred}^{(t+1)} + \frac{\gamma}{2|\mathcal{F}_\ma| + 1}$
\ENDFOR
\ENSURE Sequence of (randomized) quantile predictors ${\theta}_1, \ldots, {\theta}_T$
\end{algorithmic}
\end{algorithm}

%% file: files/5_expts.tex
\section{Experiments}
\label{sec:expts}

In this section, we present a set of empirical evaluations on real applications. In each example, we define a baseline prediction sequence $\tilde{p}_t(x_t), t \in [T]$ and a set of objectives we evaluate.~We learn locally adaptive predictions using the general recipe in Algorithm \ref{alg:fixed-share} and compare with baseline approaches we define in Section \ref{sec:baselines}. In Section~\ref{sec:datasets}, we specify, for each dataset we examine, a practically and societally meaningful set of covariates that define the function class $\gF$. Code to reproduce our experiments is available at \url{https://github.com/jivatneet/adaptive-multiobjective}.

\subsection{Datasets}
\label{sec:datasets}

\paragraph{GEFCom2014 electric load forecasting.} The Global Energy Forecasting Competition 2014 (GEFCom2014) \citep{HONG2016896} is a probabilistic energy
forecasting competition conducted with four tracks on load, price, wind and solar forecasting. In this work, we study the electricity demand forecasting track GEFCom2014-L, where participants were tasked with forecasting month-ahead quantiles of hourly load for a U.S. utility from January 1, 2011 through December 31, 2011 using historical load and temperature data. In Section~\ref{sec:peek}, we introduced the task and displayed the load and temperature trends over time in Figure~\ref{fig:load_series}. We set the function class $\gF$ to be the indicator functions for the  temperature groups $\{[0, 20),[20,40),\dots,[80,100)\}$ (in $^{\circ}$F). We consider a binary load prediction task for our empirical evaluation, in which the goal is to estimate the probability that electricity demand exceeds 150 MW during hour $t$. We construct our baseline predictions $\tilde{p}_t(x_t)$ by linearly interpolating the quantiles forecasts of~\citet{ZIEL20161029}, whose method outperforms the top entries in the competition. See Appendix~\ref{sec:gefcom_details} for details of the linear interpolation procedure.

\begin{figure}[!ht] 
    \centering
    \includegraphics[width=0.9\linewidth]{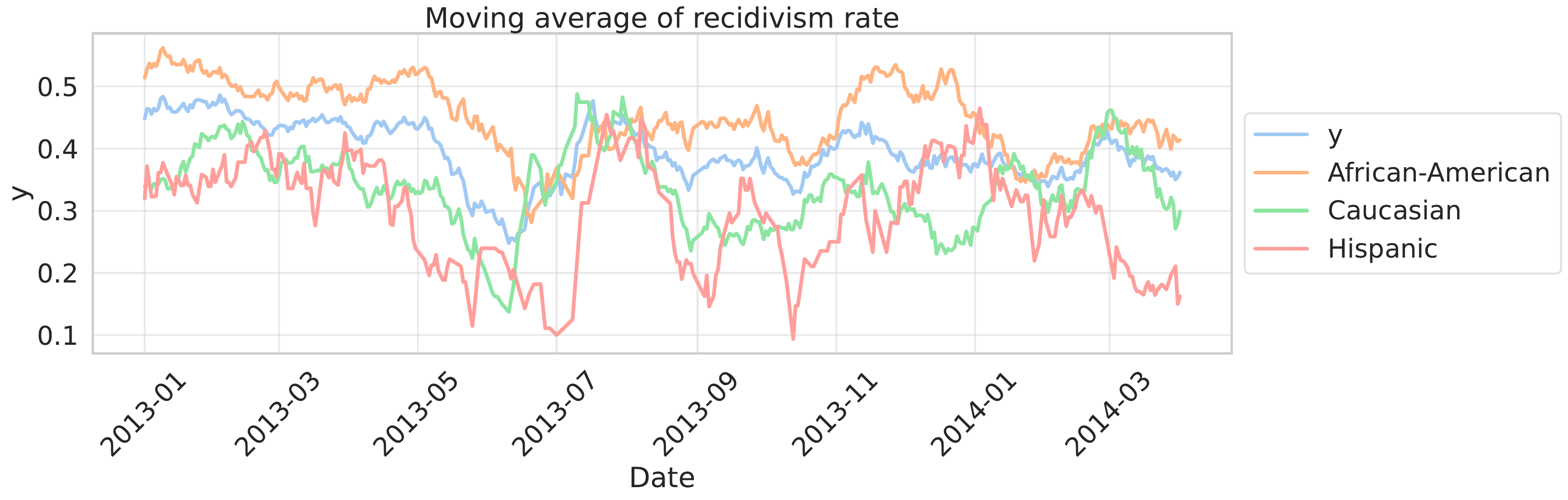} 
    \caption{\textbf{COMPAS dataset.} Moving average of true recidivism over time. We show 30-day moving averages of $y$ (recidivism indicator), computed overall and separately by racial group. For each calendar date, outcomes are first averaged across all individuals screened that day and then reported as a 30-day time-window rolling mean.}
    \label{fig:compas_y}
\end{figure}

\paragraph{COMPAS dataset.}~\citet{larson2016analyzed} analyzed the COMPAS tool used to predict recidivism for criminal defendants in Broward County, Florida and found that certain groups of defendants are more likely to be incorrectly judged as high risk of recidivism. In Figure~\ref{fig:compas_y}, we plot the true recidivism rate over time for different racial groups. We consider the recidivism prediction task and evaluate the local multiaccuracy of predictions with respect to the African-American, Caucasian, and Hispanic subgroups that constitute over 90\% of the dataset. We use the COMPAS recidivism risk scores provided in the dataset as our baseline predictions. The scores take integer values between 1–10 and we rescale to $[0,1]$ by dividing by 10. Following the analysis of~\citet{barenstein2019propublicascompasdatarevisited} who point out the data processing error in the two-year sample cutoff rule for recidivists, we drop the data points with COMPAS screen date after April 1, 2014.

\subsection{Baselines}
\label{sec:baselines}

We consider baselines that differ in their adaptivity and the set of objectives in $\gL$. \textbf{MA+pred} denotes the algorithm with the multiaccuracy and prediction error objectives $\gL := \{\ell_{\ma_{f,\sigma}} : f \in \mathcal{F}_{\text{MA}}, \sigma \in \{\pm \} \} \cup\ \{\ell_{\pred}\}$.~We explain the baselines below:
\paragraph{Baseline predictions $\tilde{p}_t(x_t)$.}These are the predictions that were constructed in advance for the application and are our input to Algorithm~\ref{alg:multiobjective-ma-regret}.

\paragraph{Multiaccuracy (MA)} with $\gL := \{\ell_{\ma_{f,\sigma}} : f \in \mathcal{F}_{\text{MA}}, \sigma \in \{\pm \} \}$: This is a specific case of Algorithm~\ref{alg:multiobjective-ma-regret} where the set $\gL$ does not include the prediction error objective.
    
\paragraph{Multicalibration (MC).} We implement the online multicalibration algorithm from~\citet{lee2022online}. This is a competitive algorithm as multicalibration is a stronger condition than multiaccuracy.~\citet{lee2022online} show that their algorithm can guarantee that predictions satisfy an accuracy objective  (specifically, low squared error) on subgroups in addition to multicalibration. Hence, we consider $c$ as the squared error in our prediction error objective. We take the number of bins as $m=10$ as it is a reasonable target for multicalibration. Lower values will give better multiaccuracy results at the cost of a much weaker multicalibration guarantee. We evaluate for varying $m$ in Appendix~\ref{sec:mc_details}. This method has an additional hyperparameter $r$ used to define a larger action space for the learner. Following \citet{lee2022online}, the value of this parameter can be arbitrarily large and we take $r= 1000$.

We consider three variants for the algorithms: \textbf{non-adaptive, locally adaptive}, and \textbf{adaptive objectives}. The non-adaptive variant corresponds to using Hedge to learn the weights in Algorithm~\ref{alg:generic_alg}; the locally adaptive variant corresponds to using the Fixed Share update as stated in Algorithm~\ref{alg:fixed-share}; and the adaptive objectives variant corresponds to using Hedge with additional objectives for all subintervals. Specifically, the adaptive objectives method augments the objectives as $\mathcal{L}_{\text{adapt.}} = \{\ell(p_t(x_t),x_t,y_t) \mathbbm{1}\{t \in I\}, \ell \in \mathcal{L}, I = [r,s] \subseteq [T]\}$.

\subsection{Local multiaccuracy and prediction error evaluation}
\label{sec:results_ma_pred}

In this section, we evaluate the local multiaccuracy error $\ell_\ma$ and prediction error $\ell_\pred$ incurred by the algorithms we defined above.

\begin{figure}[!ht] 
    \centering
    \includegraphics[width=\linewidth]{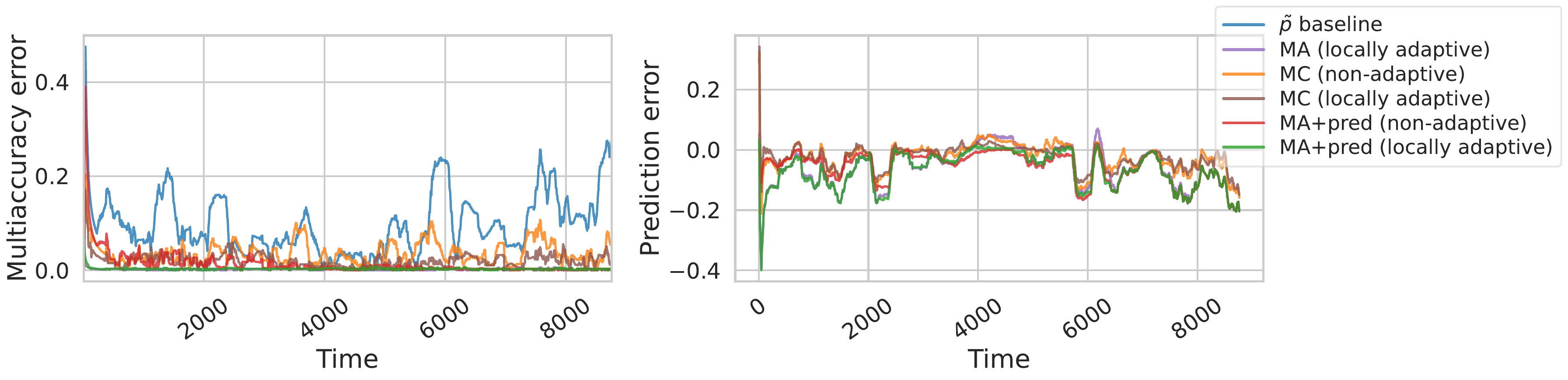} 
    \caption{Local multiaccuracy error (left) and prediction error (right) on the GEFCom2014-L dataset. We skip the first ten time steps when plotting the multiaccuracy and prediction error for improved readability.}
    \label{fig:ma_pred_load}
\end{figure}

First, we consider the results on GEFCom2014-L dataset (Figure~\ref{fig:ma_pred_load}). We take the interval width $\tau = 336$ hours (2 weeks) for this set of experiments. We show results with varying $\tau$ in Appendix~\ref{sec:varying_gamma}. We compute empirical local multiaccuracy and prediction error rates over this moving two-week window. It can be seen that the constructed baseline predictor $\tilde{p}_t$ has high local multiaccuracy error and all algorithms improve over this baseline. Overall, the locally adaptive algorithms (MA and MA+pred) have close to zero multiaccuracy error over all local intervals. On the other hand, the non-adaptive algorithms have high local variability. Notably, both the non-adaptive and the locally adaptive variants of the multicalibration algorithm (MC) have significantly slower multiaccuracy rates in practice.

Next, we turn to study the empirical local prediction error of these algorithms plotted in the right panel. As expected, the MA baseline has non-zero local prediction error and we lose accuracy with respect to the predictor $\tilde{p}_t$ in the absence of the prediction error objective. MA+pred consistently preserves or improves accuracy over $\tilde{p}_t$. As promised by the multicalibration+calibeating algorithm in~\citet{lee2022online}, we observe that MC generally has negative prediction error, although with poorer adaptivity compared to MA+pred.

\begin{figure}[!h] 
    \centering
    \includegraphics[width=\linewidth]{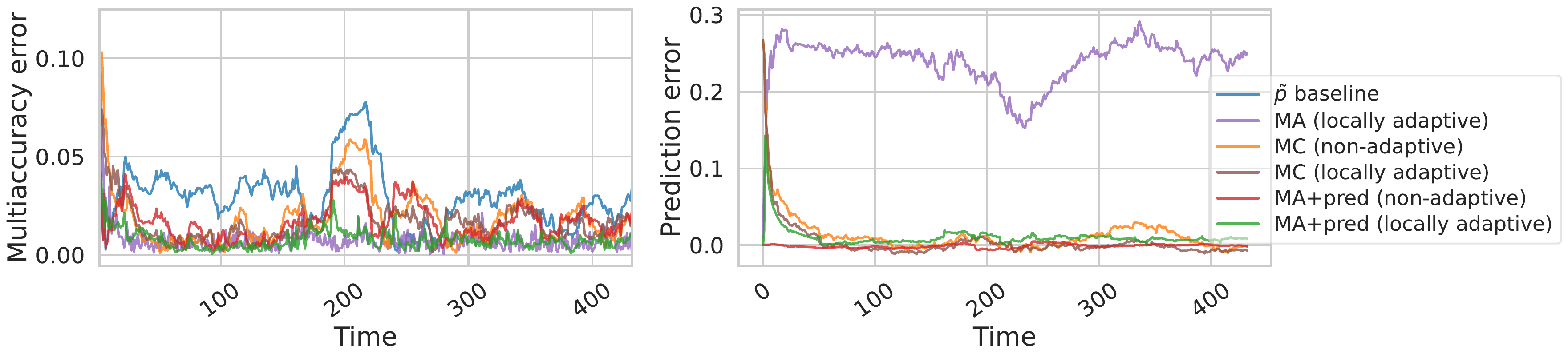} 
    \caption{Local multiaccuracy error (left) and prediction error (right) on the COMPAS dataset. We skip the first two time steps when plotting the multiaccuracy error for improved readability.}
    \label{fig:ma_pred_compas}
\end{figure}

Next, we examine our results on the COMPAS dataset (Figure~\ref{fig:ma_pred_compas}). Here, we fix $\tau=50$ days. We again see that the non-adaptive methods show minimal adaptivity to the underlying shifts and, as expected, perform poorly across some subgroups over local intervals. On the other hand, our proposed algorithm has significantly better local multiaccuracy. While the locally adaptive MC algorithm improves adaptivity relative to non-adaptive MC, its multiaccuracy rate is substantially worse than that of MA+pred (locally adaptive). Notably, it
also has higher multiaccuracy error than MA+pred (non-adaptive) on some local intervals. We note that while MA (locally adaptive) performs slightly better in terms of multiaccuracy compared to MA+pred (locally adaptive), it suffers from significantly higher prediction error over all local intervals as can be seen from the right plot in Figure~\ref{fig:ma_pred_compas}.

\subsection{Comparison with adaptive objectives multicalibration algorithm}
\label{sec:adaptive_mc_expt}

\begin{figure}[h!] 
    \centering
    \includegraphics[width=\linewidth]{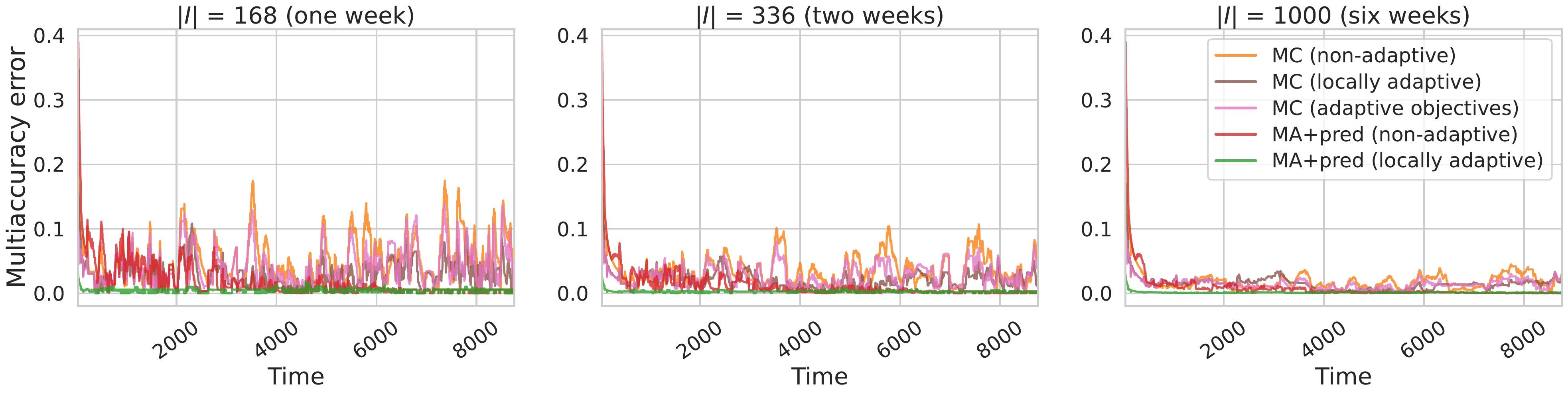} 
    \caption{Local multiaccuracy error on GEFCom2014-L for different interval widths. We skip the first thirty time steps when plotting the multiaccuracy error for improved readability.}
    \label{fig:gefcom_adaptive}
\end{figure}

\begin{figure}[h!] 
    \centering
    \includegraphics[width=\linewidth]{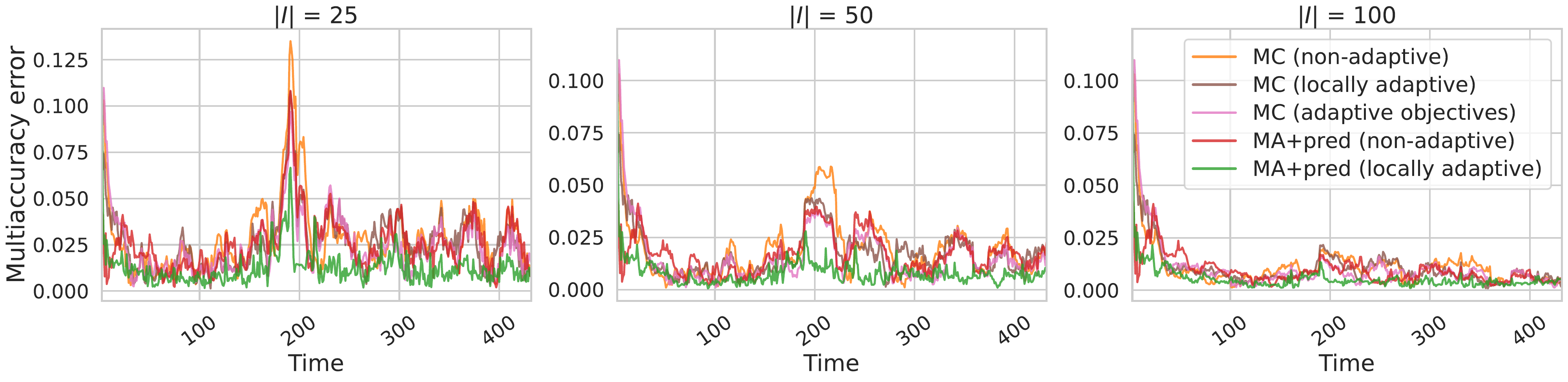} 
    \caption{Local multiaccuracy error on COMPAS for different interval widths. We skip the first two time steps when plotting the multiaccuracy error for improved readability.}
    \label{fig:compas_adaptive}
\end{figure}

Finally, we compare our algorithm with an adaptive extension of the online multicalibration algorithm proposed in~\citet{lee2022online} (MC (adaptive objectives)) in Figures~\ref{fig:gefcom_adaptive} and~\ref{fig:compas_adaptive}. This algorithm, discussed in Section~\ref{sec:algorithm}, guarantees low multicalibration error on all subintervals in $[T]$ at the expense of higher runtime and memory. While we use the fixed width values $\tau = 336$ for GEFCom2014-L and $\tau = 50$ for COMPAS in our locally adaptive  algorithm, we perform a general evaluation here over different interval widths $|I|$. We find that while adaptivity improves the performance of the multicalibration algorithm, MA+pred (locally adaptive) still has significantly better local multiaccuracy across all interval widths on both datasets. In Appendix~\ref{sec:varying_I}, we show quantitative results for a wider range of window sizes $|I|$.

\subsection{Comparison with adaptive objectives MA+pred}
\label{sec:adaptive_ma_expt}

\begin{figure}[htb!]
    \centering
    \begin{subfigure}[t]{\linewidth}
        \includegraphics[width=\linewidth]{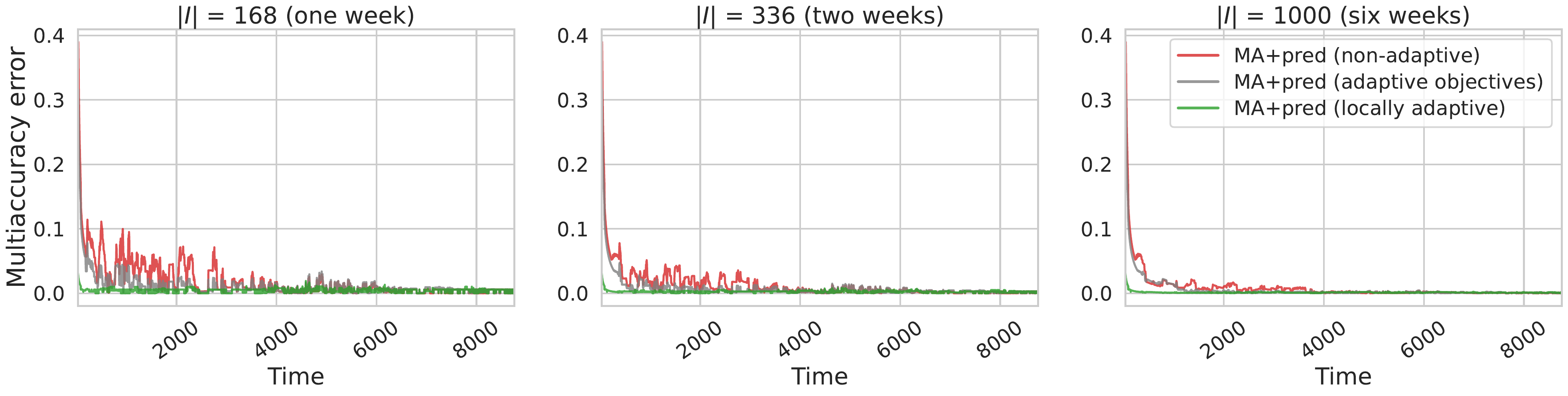}
        \caption{GEFCom2014-L}
        \label{fig:load_adaptive_marega}
    \end{subfigure}
    \par\medskip
    \begin{subfigure}[t]{\linewidth}
        \includegraphics[width=\linewidth]{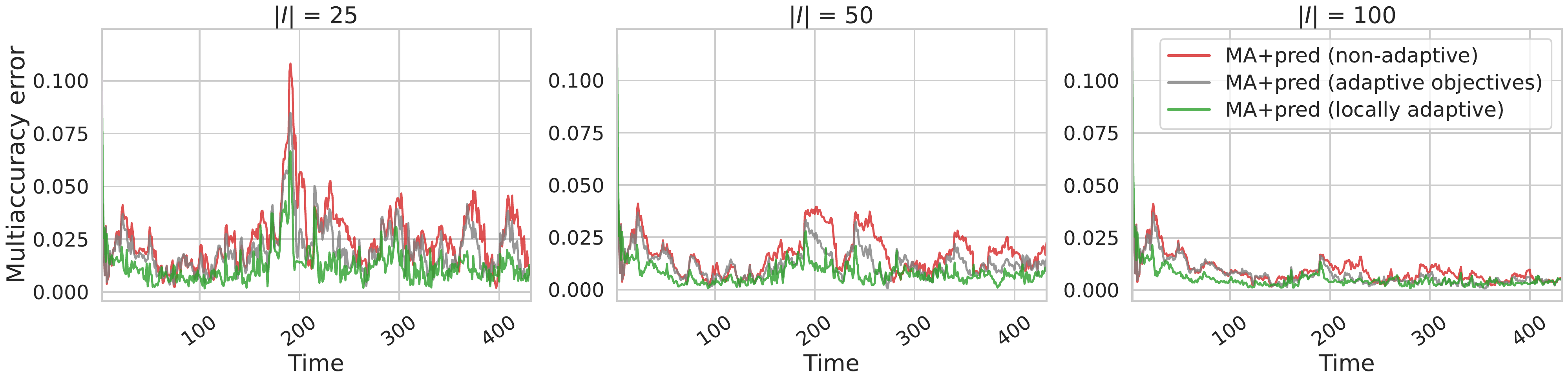}
        \caption{COMPAS}
        \label{fig:compas_adaptive_mareg}
    \end{subfigure}
   
    \caption{\textbf{Local multiaccuracy error for different interval widths $\boldsymbol{|I|}$,} (a) GEFCom2014-L and (b) COMPAS. This is the same setting as Figure \ref{fig:gefcom_adaptive} and Figure \ref{fig:compas_adaptive} where we now show comparison with the adaptive objectives MA+pred algorithm.}
    \label{fig:adaptive_mareg} 
\end{figure}

\begin{figure}[h!]
    \centering
    \begin{subfigure}{0.5\textwidth}
        \centering
        \includegraphics[width=\linewidth]{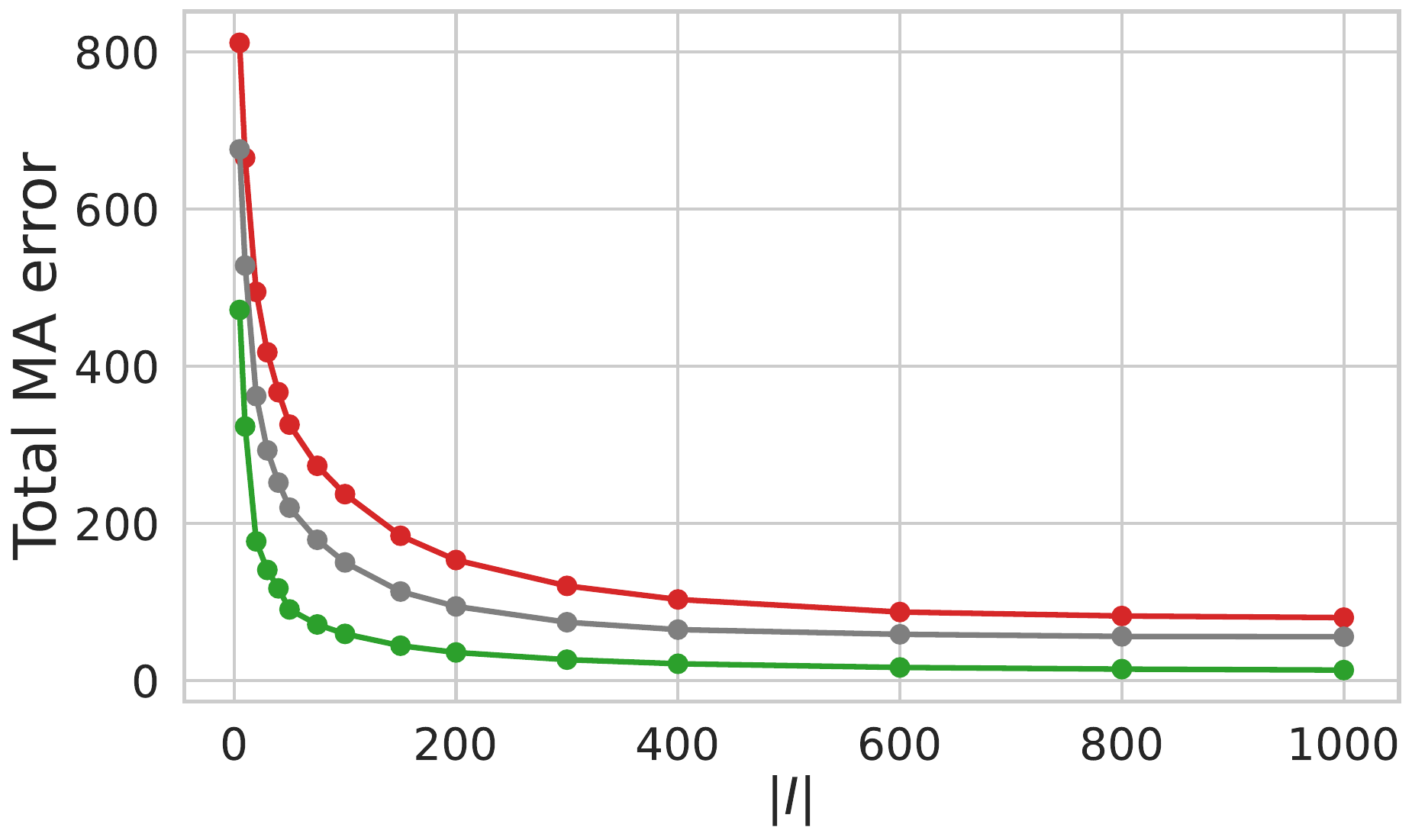}
        \caption{GEFCom2014-L}
        \label{fig:load_total_adaptive_mareg}
    \end{subfigure}
    \hfill
    \begin{subfigure}{0.46\textwidth}
        \centering
        \includegraphics[width=\linewidth]{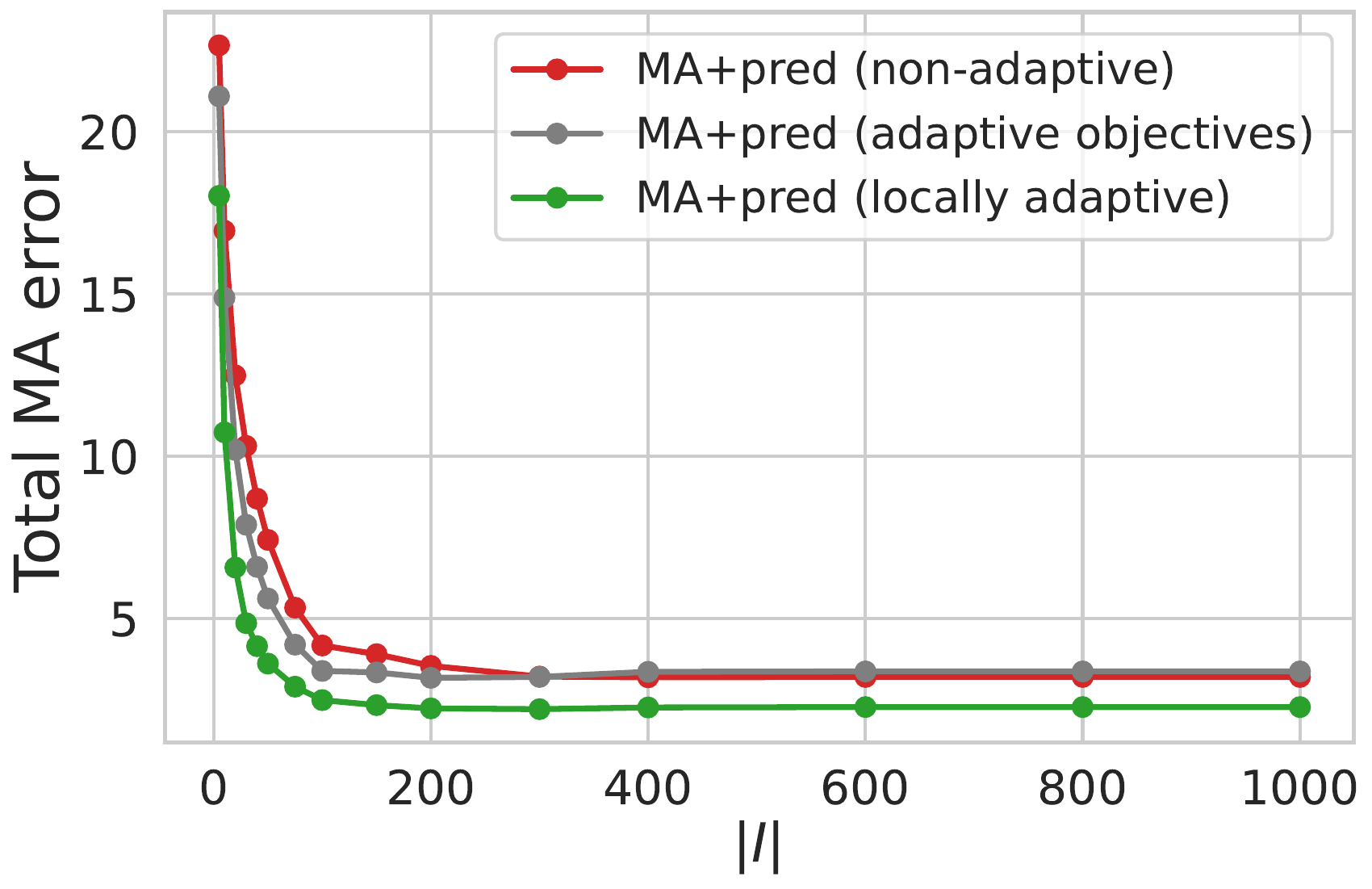}
        \caption{COMPAS}
        \label{fig:compas_total_adaptive_mareg}
    \end{subfigure}
    \caption{\textbf{Total multiaccuracy error with varying interval width $\boldsymbol{|I|}$,} (a) GEFCom2014-L and (b) COMPAS. This is the same setting as Figure \ref{fig:gefcom_adaptive} and Figure \ref{fig:compas_adaptive}. We vary the window width $|I|$ used for the moving average of errors and plot the total multiaccuracy error under the curve.}
    \label{fig:total_ma_error_adaptive_mareg}
\end{figure}

In Section~\ref{sec:adaptive_mc_expt}, we compared our proposed locally adaptive MA+pred algorithm with the adaptive online multicalibration algorithm proposed in \citet{lee2022online}. Now, we use the adaptive method proposed in \citet{lee2022online} with the MA+pred objectives. See Figures \ref{fig:adaptive_mareg} and \ref{fig:total_ma_error_adaptive_mareg} for the results, where the algorithm is labeled as MA+pred (adaptive objectives). We plot the total multiaccuracy error in Figure \ref{fig:total_ma_error_adaptive_mareg}, which is defined as the sum of the multiaccuracy errors over all local intervals of width $|I|$. Results show that while MA+pred (adaptive objectives) improves the multiaccuracy error over the non-adaptive baseline, it is consistently outperformed by MA+pred (locally adaptive) in all settings. This comparison shows that even when the adaptive baseline has the same objectives, the locally adaptive algorithm exceeds its performance.

%% file: files/6_extensions.tex
\newpage
\section{Discussion}
\label{sec:extensions}

We have presented a locally adaptive multi-objective learning algorithm that guarantees small error for all objectives over local time intervals. In this growing literature, we hope our work serves as an initial step toward bridging the empirical gap. Our evaluation focuses on a subset of multi-objective learning tasks and validation on broader problems is interesting future work. Nevertheless, many of these other problems can be readily incorporated into our framework. Table \ref{tab:extensions} provides a brief list of such problems along with their associated objectives. Further, empirical comparisons with other adaptive procedures for learning the weights could help determine whether local errors can be further reduced in practice.

\renewcommand{\arraystretch}{1.4}
\begin{table}[H]
\centering
\begin{tabular}{ccc}
\toprule
 & Objectives & Interpretation \\
\midrule
Omniprediction &  
$\ell(p_t(x_t),y_t) - \ell(f(x_t), y_t)$  & \parbox[c]{2in}{\centering\smallskip\smallskip
 $p_t$ minimizes losses $\ell \in \gL'$ against \\ competitor functions $f \in \gF$
\smallskip\smallskip}  
 \\  
\midrule
Multi-group learning & 
$\vone\{x_t \in g\}(\ell(p_t(x_t),y_t) - \ell(f(x_t), y_t))$ &   
\parbox[c]{2in}{\centering\smallskip\smallskip
 $p_t$ minimizes $\ell$ within groups $g \in \gG$ against competitor \\functions $f \in \gF$
\smallskip\smallskip} \\   
\bottomrule
\end{tabular}
\caption{Examples of extensions of our general algorithm. We define the problem, set of objectives, and the interpretation of the objectives. $\gL'$ denotes a finite class of losses.}
\label{tab:extensions}
\end{table}
\renewcommand{\arraystretch}{1}

%% file: files/ack.tex
\section*{Acknowledgements}

We thank Nika Haghtalab, Eric Zhao, and Paula Gradu for helpful discussions. We thank Florian Ziel for sharing the GEFCom2014-L quantile forecasts from their paper \citep{ZIEL20161029}. This work
was supported in part by the Office of Naval Research under grant number N00014-20-1-2787 and by the European Union (ERC-2022-SYG-OCEAN-101071601). Views and opinions expressed are however those of the author(s) only and do not
necessarily reflect those of the European Union or the European Research Council
Executive Agency. Neither the European Union nor the granting authority can be
held responsible for them.

%% file: files/app_proofs.tex
\section{Proofs}
\label{sec:proofs}

\subsection{Proof of Lemma \ref{lem:expert_interval_regret}}
We follow the calculations of \citet{pmlr-v211-gradu23a} and \citet{JMLR:v25:22-1218}. Note that while in those earlier papers the losses are nonnegative, in our work the losses may take on negative values. 

To aid in the proof we define a sequence of weights using initialization $w_{\ell}^{(t)} = 1$, for all $\ell \in \mathcal{L}$ and updates $w_{\ell}^{(t+1)} = (1 - \gamma)w_{\ell}^{(t)}  \exp\left( \eta \cdot \ell(p_t(x_t),x_t,y_t)) \right) + \gamma W^{(t+1)}/|\gL|$, where $W^{(t+1)} := \sum_{\ell} w_{\ell}^{(t)}  \exp\left( \eta \cdot \ell(p_t(x_t),x_t,y_t)) \right)$. By construction, the probabilities appearing in Algorithm \ref{alg:generic_alg} are given as  $q_{\ell}^{(t)} = \frac{w_{\ell}^{(t)}}{\sum_{\ell} w_{\ell}^{(t)}}$. Thus, 

\[
    \frac{W^{(t+1)}}{W^{(t)}} = \sum_{\ell \in \gL} q_{\ell}^{(t)}  \exp\left( \eta \cdot \ell(p_t(x_t),x_t,y_t)) \right).
\]
Since $\eta \leq 1$ and $\ell$ is bounded between $[-1,1]$, $|\eta \cdot \ell(p_t(x_t),x_t,y_t)| \leq 1$. We use the inequalities $1 + a \leq \exp(a)$ and for $|a| \leq 1, \exp(a) \leq 1 + a + a^2$ to get

\[
\frac{W^{(t+1)}}{W^{(t)}} \leq \exp{( \eta q^{(t)\top} \ell^{(t)} + \eta^2 q^{(t)\top} {\ell^{(t)}_\gL}^2)}.
\]
Inductively, this implies that, for any interval $I = [r, s]$,

\[
    \frac{W^{(s+1)}}{W^{(r)}} \leq \exp \Bigg(\sum_{t=r}^s  \eta q^{(t)\top} \ell^{(t)} + \eta^2 q^{(t)\top} {\ell^{(t)}_\gL}^2 \Bigg).
\]
On the other hand, for any fixed $\ell \in \gL$, $w_{\ell}^{(t+1)} \geq w_{\ell}^{(t)}(1 - \gamma)\exp{(\eta \ell^{(t)})}$. Without loss of generality, we proceed with a fixed $\ell$, noting that the same calculations will follow for all $\ell \in \gL$. This gives
\begin{align*}
    \frac{W^{(s+1)}}{W^{(r)}} \geq \frac{w_{\ell}^{(s+1)}}{W^{(r)}} &\geq (1 - \gamma)^{|I|} w_{\ell}^{(r)} \exp \Bigg(\sum_{t=r}^s  \eta \ell^{(t)} \Bigg) \\
    &\geq (1 - \gamma)^{|I|} \frac{\gamma}{|\gL|} \exp \Bigg(\sum_{t=r}^s  \eta \ell^{(t)} \Bigg). 
\end{align*}
Combining the two inequalities and taking logarithm on both sides yields
\begin{align*}
    |I|\log(1 - \gamma) + \log \bigg( \frac{\gamma}{|\gL|}\bigg) + \sum_{t=r}^s  \eta \ell^{(t)} \leq \sum_{t=r}^s  \eta q^{(t)\top} \ell^{(t)}_\gL + \eta^2 q^{(t)\top} {\ell^{(t)}_\gL}^2.
\end{align*}
We rearrange to get the following inequality
\begin{align*}
    \sum_{t=r}^s  q^{(t)\top} \ell^{(t)}_\gL \geq \sum_{t=r}^s \ell^{(t)} - \eta \Bigg(\sum_{t=r}^s q^{(t)\top} {\ell^{(t)}_\gL}^2 \Bigg) + \frac{1}{\eta} |I|\log(1 - \gamma) + \log \bigg( \frac{\gamma}{|\mathcal{L}|}\bigg).
\end{align*}
As $\gamma \leq 1/2$, we can use the inequality $\log(1-\gamma) \geq -2\gamma$ to get the final inequality
\begin{align*}
     \sum_{t=r}^s q^{(t)\top} \ell^{(t)}_\gL \geq \sum_{t=r}^{s} \ell^{(t)} - \eta \Bigg(\sum_{t=r}^s q^{(t)\top} {\ell^{(t)}_\gL}^2 \Bigg) - \frac{1}{\eta}\bigg(\log\bigg(\frac{|\mathcal{L}|}{\gamma} \bigg) + |I|2\gamma \bigg).
\end{align*}
As the same calculation holds for any objective $\ell \in \gL$, we get the final result
\begin{equation*}
    \hspace{-3mm}
     \sum_{t=r}^s  q^{(t)\top} \ell^{(t)}_\gL \geq  \underset{\ell \in \gL}{\operatorname{max}} \sum_{t=r}^s \ell^{(t)} - \eta \Bigg(\sum_{t=r}^s q^{(t)\top} {\ell^{(t)}_\gL}^2 \Bigg) - \frac{1}{\eta}\bigg(\log\bigg(\frac{|\mathcal{L}|}{\gamma} \bigg) + |I|2\gamma \bigg).
\end{equation*}

\subsection{Proof of Lemma \ref{lem:expected_l_le_0}}

This result was shown in~\citet{lee2022online} and we include their argument here for completeness.

Let $u^{(t)}(p, y) := \sum\limits_{\ell} q_{\ell}^{(t)} \ell(p, x_t, y).$ Let $\Delta(\mathcal{Y})$ denote the space of distributions over $\mathcal{Y}$. Applying Sion's Minimax Theorem, we get 
\[
\underset{P \in \Delta(\mathcal{Y})}{\operatorname{min}} \;\underset{y \in \gY}{\operatorname{max}}\;\mathbb{E}_{p \sim P}[u^{(t)}(p, y)] = \underset{P \in \Delta(\mathcal{Y})}{\operatorname{min}} \;\underset{Q \in \Delta(\gY)}{\operatorname{max}}\;\mathbb{E}_{p \sim P ,y \sim Q}[u^{(t)}(p, y)]  = \underset{Q \in \Delta(\gY)}{\operatorname{max}}\;\underset{P \in \Delta(\gY)}{\operatorname{min}} \;\mathbb{E}_{p \sim P ,y \sim Q}[u^{(t)}(p, y)].  
\] 
This conveys that the minimax-optimal strategy $p_t$ of the learner can achieve $u^{(t)}(p, y)$ as low as if the adversary moved first and the learner could best-respond. Now, for a fixed distribution $Q$ on $\gY$ we have that by Assumption \ref{assump:consistent_objs} there exists $p^*$ such that $\mathbb{E}_{y \sim Q} [u^{(t)}(p^*, y)] \leq 0$.

Thus, the minimax optimal strategy guarantees that $\min_{P \in \Delta(\gY)} \max_{y \in \mathcal{Y}} \mathbb{E}_{p \sim P}[u^{(t)}(p, y)] \leq 0$ for all $t \in [T]$. This yields the desired inequality
\[
    \sum_{t=r}^s q^{(t)\top} \ell^{(t)}_\gL \leq 0.
\]

\subsection{Proof of Theorem \ref{thm:adaptive_loss_bound}}

Applying Lemma~\ref{lem:expected_l_le_0} to the inequality (\ref{eq:lem1_mo}) in Lemma \ref{lem:expert_interval_regret} gives
\begin{equation*}
    \underset{\ell \in \gL}{\operatorname{max}} \sum_{t=r}^s \ell^{(t)} - \eta \Bigg(\sum_{t=r}^s q^{(t)\top} {\ell^{(t)}_\gL}^2\Bigg) - \frac{1}{\eta}\bigg(\log\bigg(\frac{|\mathcal{L}|}{\gamma} \bigg) + |I|2\gamma \bigg) \leq 0.
\end{equation*}
Rearranging and dividing both sides by $|I|$ yields the desired inequality,
\begin{equation*}
    \underset{\ell \in \gL}{\operatorname{max}} \frac{1}{|I|} \sum_{t=r}^s \ell^{(t)} \leq \frac{\eta}{|I|} \Bigg(\sum_{t=r}^s q^{(t)\top} {\ell^{(t)}_\gL}^2\Bigg) + \frac{1}{\eta |I|}\bigg(\log\bigg(\frac{|\mathcal{L}|}{\gamma} \bigg) + |I|2\gamma \bigg).
\end{equation*}

\subsection{Proof of Corollary \ref{cor:multiaccuracy}}

The proof follows by instantiating the set of objectives $\gL$ for multiaccurate mean estimation in Theorem~\ref{thm:adaptive_loss_bound}. We take $\gL := \{\ell_{\ma_{f,\sigma}} : f \in \mathcal{F}_{\text{MA}}, \sigma \in \{\pm \} \} \cup\ \{\ell_{\pred}\}$, where $\gF_{\ma} \subseteq \{f: \gX \rightarrow [0,1] \}$ is the function class that we desire multiaccuracy with respect to and $\ell_{\pred}$ is the prediction error objective. Plugging the objectives in \eqref{eq:mo_final_upper_bound}, this gives us the desired bound.

\subsection{Proof of Corollary \ref{cor:multiaccuracy_quantiles}}

The proof follows by instantiating the set of objectives $\gL$ for multiaccurate quantile estimation in Theorem~\ref{thm:adaptive_loss_bound}. We take $\gL := \{\sigma f(x_t)(\vone \{y_t \leq \theta_t\} - \alpha) : f \in \mathcal{F}_{\text{MA}}, \sigma \in \{\pm \} \} \cup\ \{\ell_\alpha(\theta_t, y_t) - \ell_\alpha(\tilde{\theta}_t, y_t)\}$, where $\gF_{\ma} \subseteq \{f: \gX \rightarrow [0,1] \}$ is the function class that we desire multiaccuracy with respect to. Plugging the objectives in \eqref{eq:mo_final_upper_bound}, this gives us the desired bound.

\section{Deferred Algorithms}
\label{sec:app_extensions}

In Section \ref{sec:acc_objective_significance}, we discussed the significance of the prediction error objective in preserving the accuracy relative to a base predictor sequence $\tilde{p}_t(x_t)$. When $\tilde{p}_t(x_t)$ is not available in advance, we can combine our procedure with a standard online learning algorithm (e.g., online gradient or mirror descent) that provides an appropriate baseline. Algorithm~\ref{alg:multiobjective-ma-regret-nobaseline} gives a complete description of this approach. In what follows, the weights $q^{(t)}_{\text{MA},f,\sigma}$ and $q_{\pred}^{(t)}$ are used to denote the entries of $q^{(t)}$ associated with the multiaccuracy and prediction error objectives, respectively.

\begin{algorithm}[h]
\caption{Locally adaptive multiaccurate mean estimation (learning $\tilde{p}_t$ online)}
\label{alg:multiobjective-ma-regret-nobaseline}
\begin{algorithmic}[1]
\REQUIRE Function class $\gF_{\ma} \subseteq \{f: \gX \rightarrow [0,1] \}$; $\gF_{\pred} = \{f_\beta : \beta \in \sR\}$; hyperparameters $\eta, \gamma, \zeta$. 
\REQUIRE Sequence of samples $\{ (x_1, y_1), \ldots, (x_T, y_T) \}$ 
\STATE $q_{{\ma}_{f, \sigma}}^{(1)} =\frac{1}{2|\mathcal{F}_\ma| + 1}, \quad \forall f \in \mathcal{F}_{\ma}, \sigma \in \{\pm 1\}$.
\STATE $q_{\pred}^{(1)} =\frac{1}{2|\mathcal{F}_\ma| + 1}$
\STATE $\beta_1 = 0$
\FOR{each $t \in [T]$}
    \STATE $\tilde{p}_t(x_t) := f_{\beta_t}(x_t)$
    \STATE $p_t(x_t) :=  \underset{p \in \mathcal{Y}}{\operatorname{argmin}} \;\underset{y \in \gY}{\operatorname{max}}\; \sum\limits_{f,\sigma} q_{\ma_{f,\sigma}}^{(t)} \sigma f(x_t)(y - p) + q_{\pred}^{(t)} (c(p, y) - c(\tilde{p}_t(x_t), y))$
    \STATE $\tilde{q}_{\ma_{f,\sigma}}^{(t+1)} = \dfrac{q_{\ma_{f,\sigma}}^{(t)} \exp\left( \eta \cdot \sigma f(x_t)(y_t - p_t(x_t)) \right)}{\sum_{\ell' \in \mathcal{L}} q_{\ell'}^{(t)} \exp\left( \eta \cdot \ell'(p_t(x_t),x_t,y_t) \right)}$ for all $f \in \mathcal{F}_\ma, \sigma \in \{\pm 1\}$
    \STATE $\tilde{q}_{\pred}^{(t+1)} = \dfrac{q_{\pred}^{(t)} \exp\left( \eta \cdot (c(p_t(x_t), y_t) - c(\tilde{p}_t(x_t), y_t)  )\right)}{\sum_{\ell' \in \mathcal{L}} q_{\ell'}^{(t)} \exp\left( \eta \cdot \ell'(p_t(x_t),x_t,y_t) \right)}$
    \STATE $q_{\ma_{f,\sigma}}^{(t+1)} = (1 - \gamma) \, \tilde{q}_{\ma_{f,\sigma}}^{(t+1)} + \frac{\gamma}{2|\mathcal{F}_\ma| + 1}$
    \STATE $q_{\pred}^{(t+1)} = (1 - \gamma) \, \tilde{q}_{\pred}^{(t+1)} + \frac{\gamma}{2|\mathcal{F}_\ma| + 1}$
    \STATE $\beta_{t+1} = \beta_t - \zeta \nabla_{\beta} \ell \left(f_{\beta_t}(x_t), y_t \right)$
    \label{line:finite-ma-exp}
\ENDFOR
\ENSURE Sequence of predictions $p_1(x_1), \ldots, p_T(x_T)$
\end{algorithmic}
\end{algorithm}

%% file: files/app_expt_details.tex
\section{Additional Experimental Details}
\label{sec:expt_details}

\subsection{GEFCom2014 electric load forecasting}
\label{sec:gefcom_details}

For our electric load forecasting experiment, we need to compute the hourly probability that electricity demand exceeds a threshold (150 MW in our example) given quantile forecasts. We use linear interpolation to estimate the full
cumulative distribution function of the load from the quantile forecasts of~\citet{ZIEL20161029}. Their method outperforms top entries of the competition. Fix a set of quantile levels $0 < \alpha_1 < \dots < \alpha_k$ and let the corresponding set of quantile forecasts at hour $t$ be $\hat{\theta}_t^{\alpha_1} < \dots < \hat{\theta}_t^{\alpha_k}$. Let $Y_t \in \sR$ denote the hourly load. We estimate the cumulative distribution function of $Y_t$ by linearly interpolating between the points $\{(\hat{\theta}_t^{\alpha_i}, \alpha_i)\}_{i=1}^k$. Formally, for any $x \in \sR$

\[
\widehat{\mathbb{P}}(Y \le x) =
\begin{cases}
    0, & x < \alpha_1,\\
    1, & x \ge \alpha_k,\\
    \alpha_{i-1} + \dfrac{\alpha_i - \alpha_{i-1}}{\hat{\theta}_t^{\alpha_{i}}- \hat{\theta}_t^{\alpha_{i-1}}}\,\bigl(x - \hat{\theta}_t^{\alpha_{i-1}}\bigr),
    & \hat{\theta}_t^{\alpha_{i-1}} \le x < \hat{\theta}_t^{\alpha_{i}}.
\end{cases}
\]

\begin{figure}[h!] 
    \centering
    \includegraphics[width=0.7\linewidth]{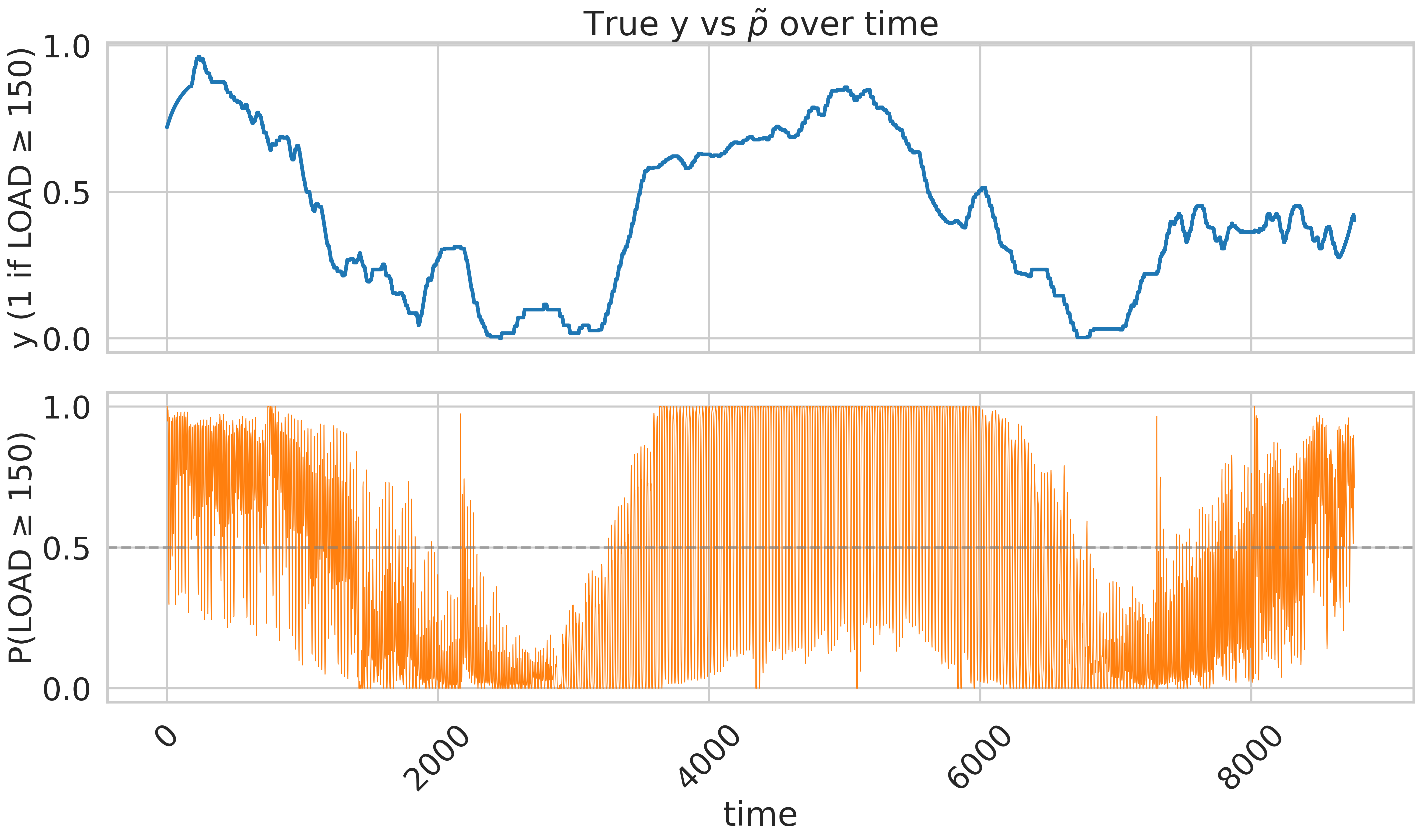} 
    \caption{\textbf{GEFCom2014-L: True load ($y$) and constructed predictions $\tilde{p}_t$}. We plot the moving average of the binary $y$ over a window size $|I| =$ 336 hours (2 weeks) (top) and the baselines predictions $\tilde{p}_t$ constructed from the quantile forecasts using linear interpolation (bottom) over time.}
    \label{fig:ptilde}
\end{figure}

Figure \ref{fig:ptilde} shows the constructed baseline predictions $\tilde{p}_t$ for the task using the above procedure along with the raw load variable.

\subsection{Multicalibration implementation}
\label{sec:mc_details}

\begin{figure}[htb!]
    \centering
    \begin{subfigure}[t]{0.95\linewidth}
        \includegraphics[width=\linewidth]{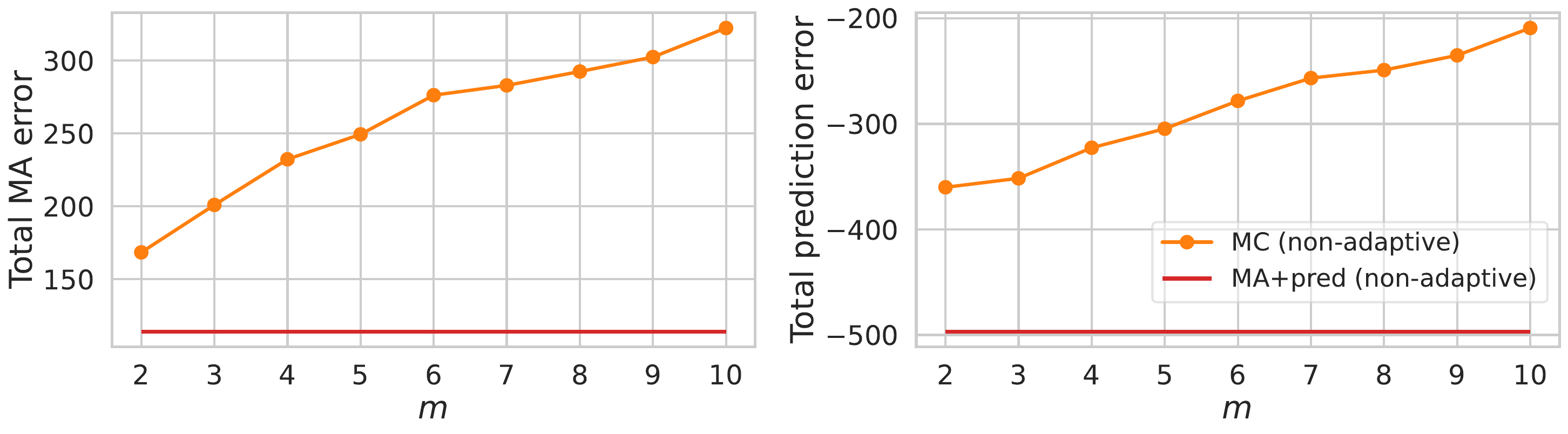}
        \caption{GEFCom2014-L}
        \label{fig:load_m_small}
    \end{subfigure}
    \par\medskip
    \begin{subfigure}[t]{0.95\linewidth}
        \includegraphics[width=\linewidth]{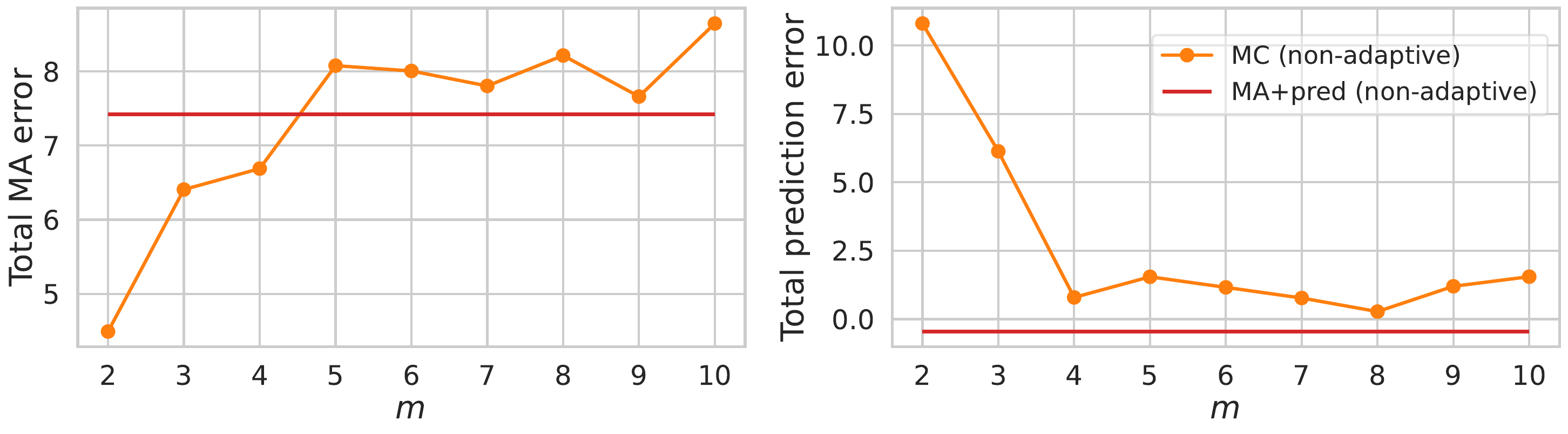}
        \caption{COMPAS}
        \label{fig:compas_m_small}
    \end{subfigure}
   
    \caption{\textbf{Total multiaccuracy error and prediction error with varying $\boldsymbol{m}$,} (a) GEFCom2014-L and (b) COMPAS. This is the same setting as Figure \ref{fig:ma_pred_load} and Figure \ref{fig:ma_pred_compas} where we now vary the number of bins $m$.}
    \label{fig:total_ma_varying_m_small} 
\end{figure}

We implement the multicalibration + calibeating algorithm in \citet{lee2022online} in order to calibeat the baseline forecaster sequence $\tilde{p}_t(x_t)$. We set the optimal choice of $\eta = \sqrt{\frac{\log(2|\gL|m)}{4T}}$ that minimizes the regret bound for the algorithm. We take the number of bins $m=10$ and use 10 level sets of the forecaster throughout. Figure~\ref{fig:total_ma_varying_m_small} shows the total multiaccuracy error and prediction error for varying values of $m$. The total multiaccuracy error and prediction error are defined as the sum of the multiaccuracy and prediction errors respectively over all local intervals of width $\tau$. As $m$ decreases, the multicalibration algorithm approaches the multiaccuracy algorithm and the total MA error decreases. Nevertheless, even when $m=2$, MA+pred has lower total MA error and prediction error than MC on GEFCom2014-L (Figure \ref{fig:load_m_small}). While the total MA error of MC drops below MA+pred on COMPAS with smaller $m$ (Figure \ref{fig:compas_m_small}), this is accompanied by an increase in total prediction error.

%% file: files/app_further_expts.tex
\section{Ablations on Hyperparameters}
\label{sec:ablations}

\subsection{Varying interval width $|I|$}
\label{sec:varying_I}

\begin{figure}[h!]
    \centering
    \begin{subfigure}{0.5\textwidth}
        \centering
        \includegraphics[width=\linewidth]{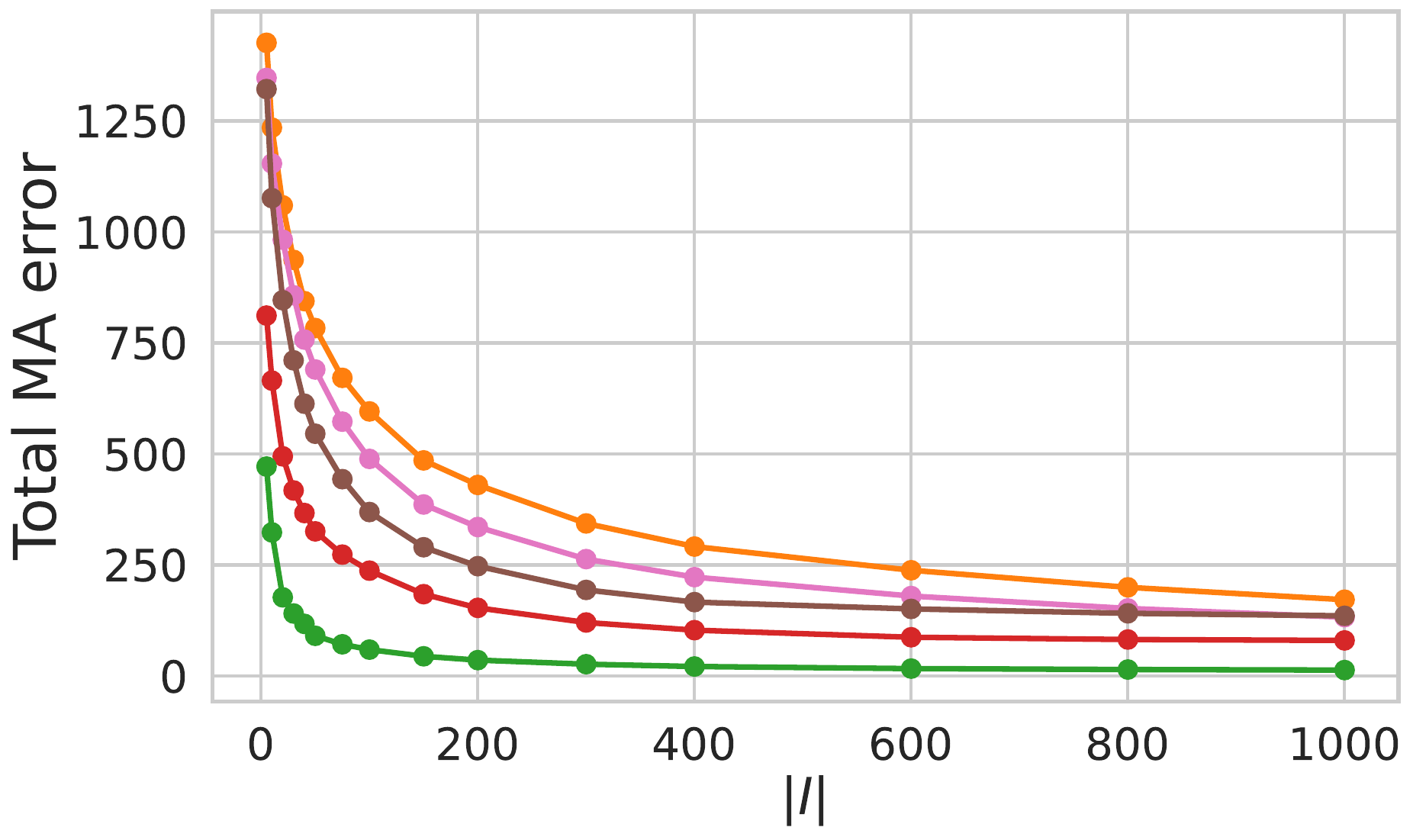}
        \caption{GEFCom2014-L}
        \label{fig:load_ma_I}
    \end{subfigure}
    \hfill
    \begin{subfigure}{0.46\textwidth}
        \centering
        \includegraphics[width=\linewidth]{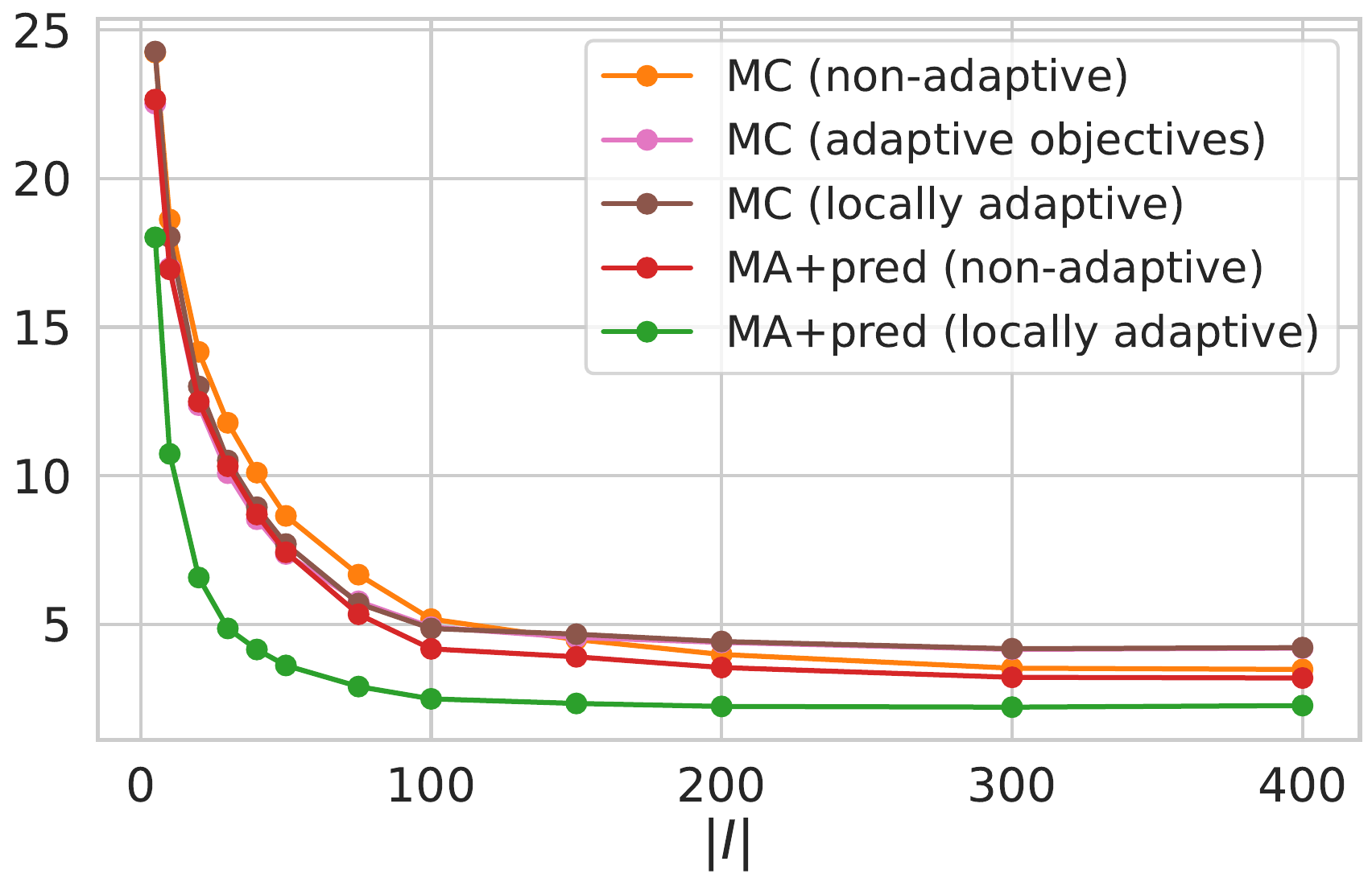}
        \caption{COMPAS}
        \label{fig:compas_ma_I}
    \end{subfigure}
   
    \caption{\textbf{Total multiaccuracy error with varying interval width $\boldsymbol{|I|}$,} (a) GEFCom2014-L and (b) COMPAS. This is the same setting as Figure \ref{fig:gefcom_adaptive} and Figure \ref{fig:compas_adaptive}. We vary the window width $|I|$ used for the moving average of errors and plot the total multiaccuracy error under the curve.}
    \label{fig:total_ma_vs_I}
\end{figure}

We extend the analysis in Section~\ref{sec:adaptive_mc_expt} and plot the total multiaccuracy error over all windows for a wide range of varying interval widths $|I|$. Results in Figure \ref{fig:total_ma_vs_I} show that locally adaptive MA+pred consistently outperforms all other adaptive algorithms despite being tuned with a fixed width. While MC (locally adaptive) improves upon the non-adaptive MC algorithm, the multiaccuracy error remains significantly higher than MA+pred (locally adaptive). It is interesting to note that MC (adaptive objectives) does not achieve lower total multiaccuracy error than MC (locally adaptive) despite its stronger theoretical guarantee over all subintervals.

\subsection{Varying $\eta$}
\label{sec:varying_eta}

\begin{figure}[htb!]
    \centering
    \begin{subfigure}[t]{\linewidth}
        \includegraphics[width=\linewidth]{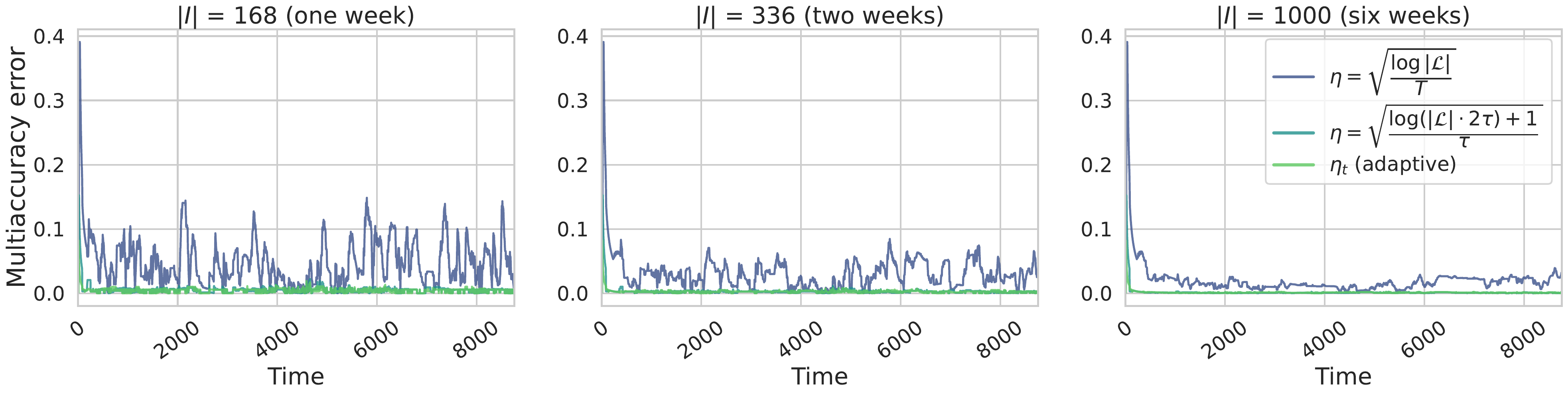}
        \caption{GEFCom2014-L}
        \label{fig:load_varying_eta}
    \end{subfigure}
    \par\medskip
    \begin{subfigure}[t]{\linewidth}
        \includegraphics[width=\linewidth]{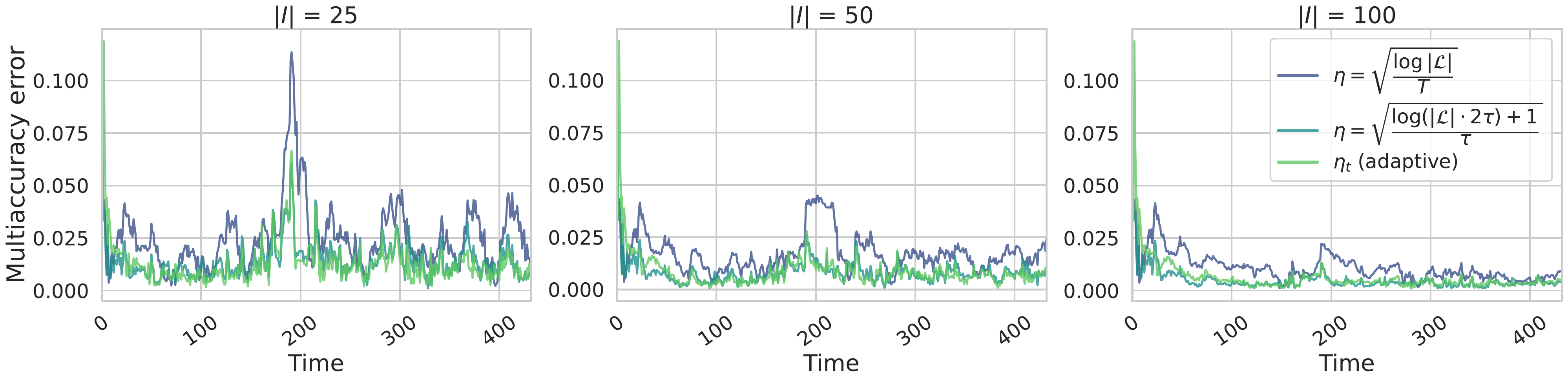}
        \caption{COMPAS}
        \label{fig:compas_varying_eta}
    \end{subfigure}
   
    \caption{\textbf{Local multiaccuracy error with varying $\boldsymbol{\eta}$ for different interval widths $\boldsymbol{|I|}$,} (a) GEFCom2014-L and (b) COMPAS. This is the same setting as Figure \ref{fig:gefcom_adaptive} and Figure \ref{fig:compas_adaptive} where we now show results with different choices of $\eta$ in the locally adaptive MA+pred algorithm.}
    \label{fig:varying_eta} 
\end{figure}

\begin{figure}[h!]
    \centering
    \begin{subfigure}{0.5\textwidth}
        \centering
        \includegraphics[width=\linewidth]{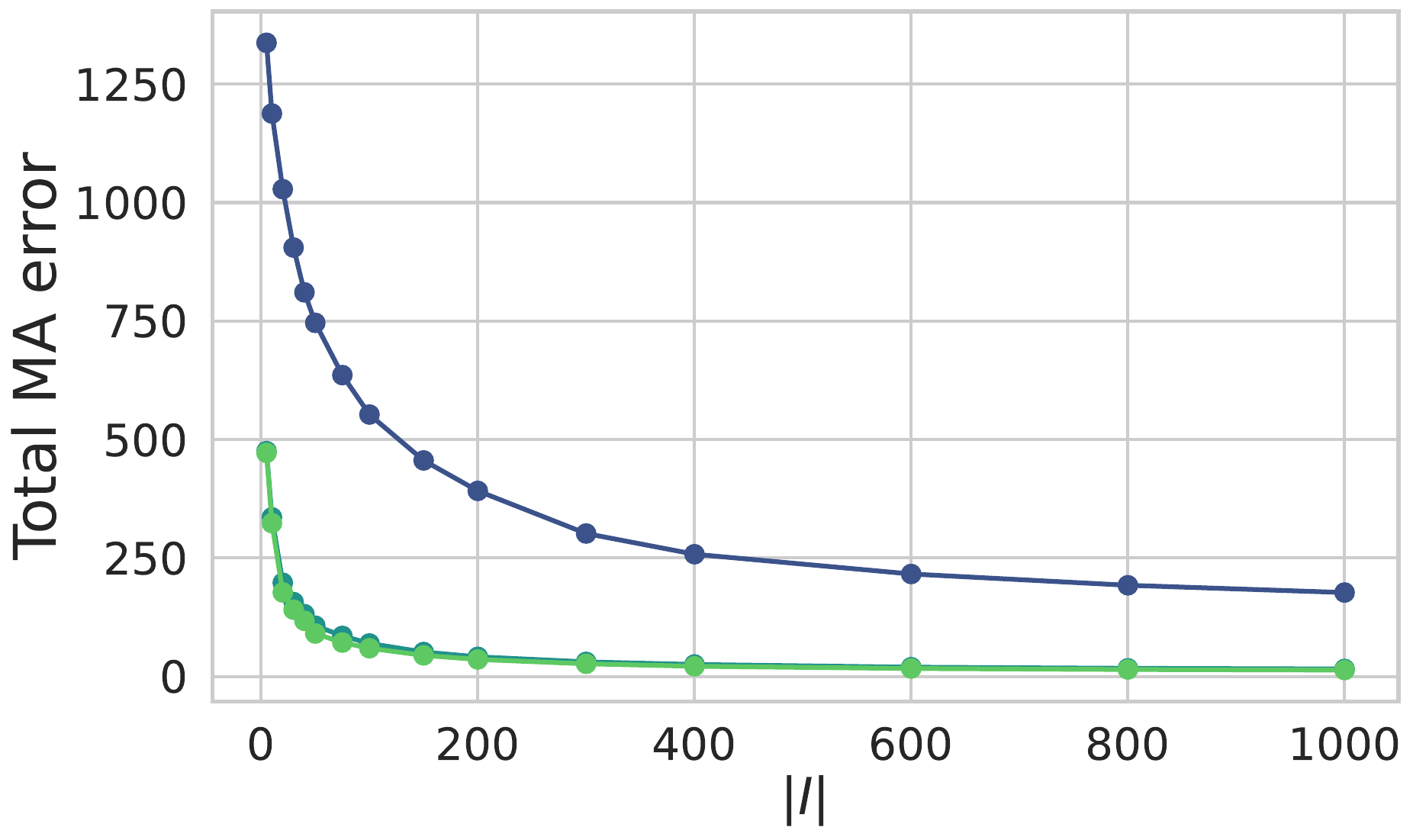}
        \caption{GEFCom2014-L}
        \label{fig:load_varying_eta_and_window}
    \end{subfigure}
    \hfill
    \begin{subfigure}{0.46\textwidth}
        \centering
        \includegraphics[width=\linewidth]{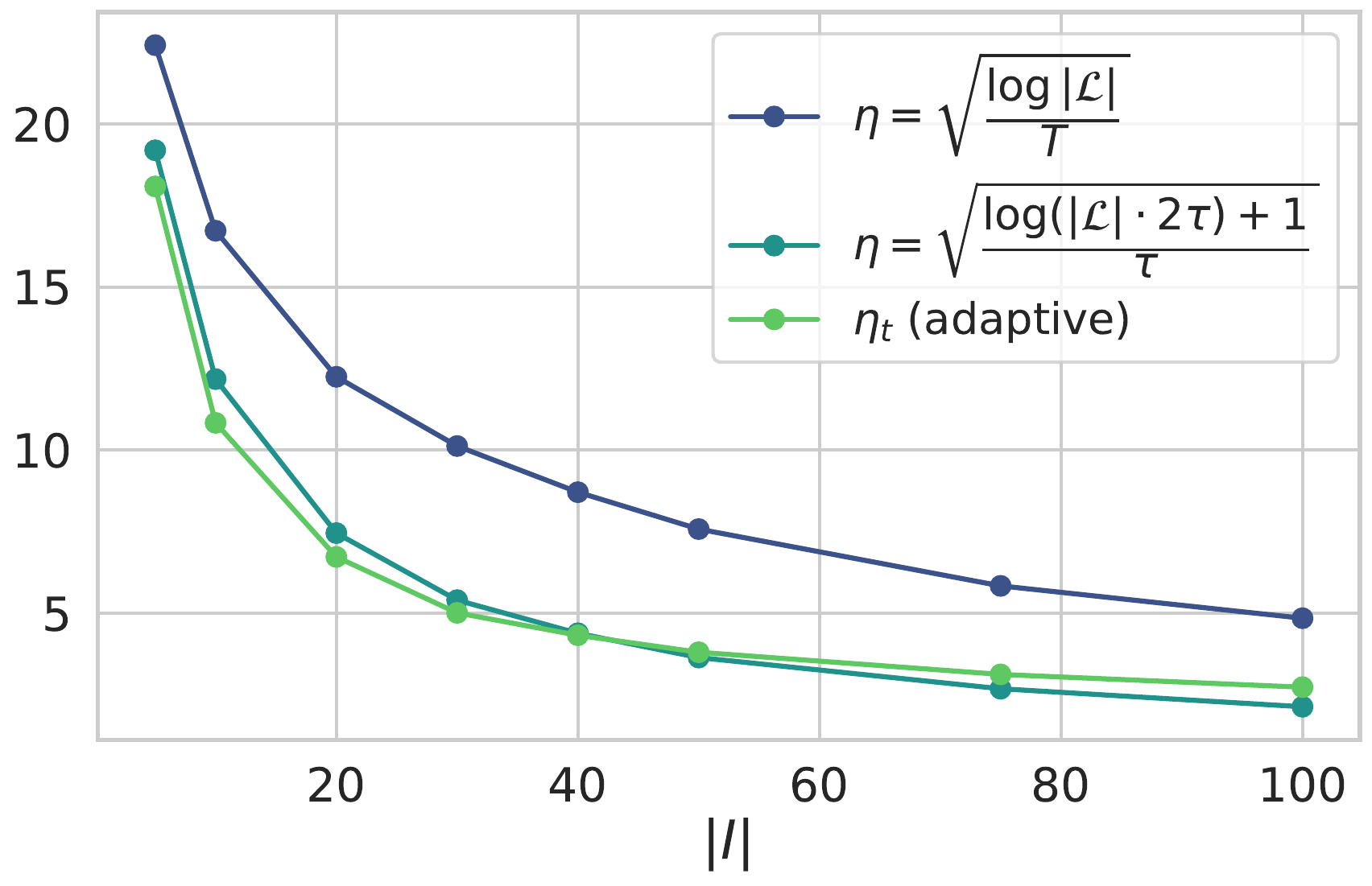}
        \caption{COMPAS}
        \label{fig:compas_varying_eta_and_window}
    \end{subfigure}
   
    \caption{\textbf{Total multiaccuracy error with varying $\boldsymbol{\eta}$ and interval width $\boldsymbol{|I|}$,} (a) GEFCom2014-L and (b) COMPAS. This is the same setting as Figure \ref{fig:gefcom_adaptive} and Figure \ref{fig:compas_adaptive}. We vary the window width $|I|$ used for the moving average of errors and plot the total multiaccuracy error under the curve with different choices of $\eta$ in the locally adaptive MA+pred algorithm.}
    \label{fig:varying_eta_and_window}
\end{figure}

In this section, we consider three different choices of $\eta$ in the locally adaptive MA+pred algorithm. 
\begin{enumerate}
    \item $\eta = \sqrt{\dfrac{\log|\gL|}{T}}$: this is the optimal $\eta$ that minimizes non-adaptive regret bound.
    \item $\eta = \sqrt{\dfrac{\log((2|\mathcal{F}_\nma| + 1)\cdot 2\tau) + 1}{\tau}}$: we substitute the online updates $\sum_{s=t-\tau +1}^{t} q_{\nma}^{(s)\top}{\ell_{\nma}^{(s)}}^2 + q_{\npredt}^{(s)} {\ell_{\npredt}^{(s)}}^2$ in the adaptive choice of $\eta_t$ (\ref{eq:adaptive_eta}) with the interval width $\tau$.
    \item $\eta = \eta_t := \sqrt{\dfrac{\log((2|\mathcal{F}_\nma| + 1)\cdot 2\tau) + 1}{\sum_{s=t-\tau +1}^{t} q_{\nma}^{(s)\top}{\ell_{\nma}^{(s)}}^2 + q_{\npredt}^{(s)} {\ell_{\npredt}^{(s)}}^2}}$: this is the adaptive choice of $\eta$ proposed in (\ref{eq:adaptive_eta}), which is the default for our algorithm. 
\end{enumerate}
Figures \ref{fig:varying_eta} and \ref{fig:varying_eta_and_window} show the local multiaccuracy error and total multiaccuracy error respectively with the above choices of $\eta$ and varying interval widths. We find that adaptive $\eta_t$ and the choice of $\eta$ that uses interval width $\tau$ consistently dominate the optimal $\eta$ for the non-adaptive regret bound. These results establish the importance of the choice of $\eta$ in achieving local adaptivity separate from the uniform exploration.

\subsection{Varying $\tau$ and $\gamma$}
\label{sec:varying_gamma}

\begin{figure}[h!]
    \centering
    \begin{subfigure}{0.5\textwidth}
        \centering
        \includegraphics[width=\linewidth]{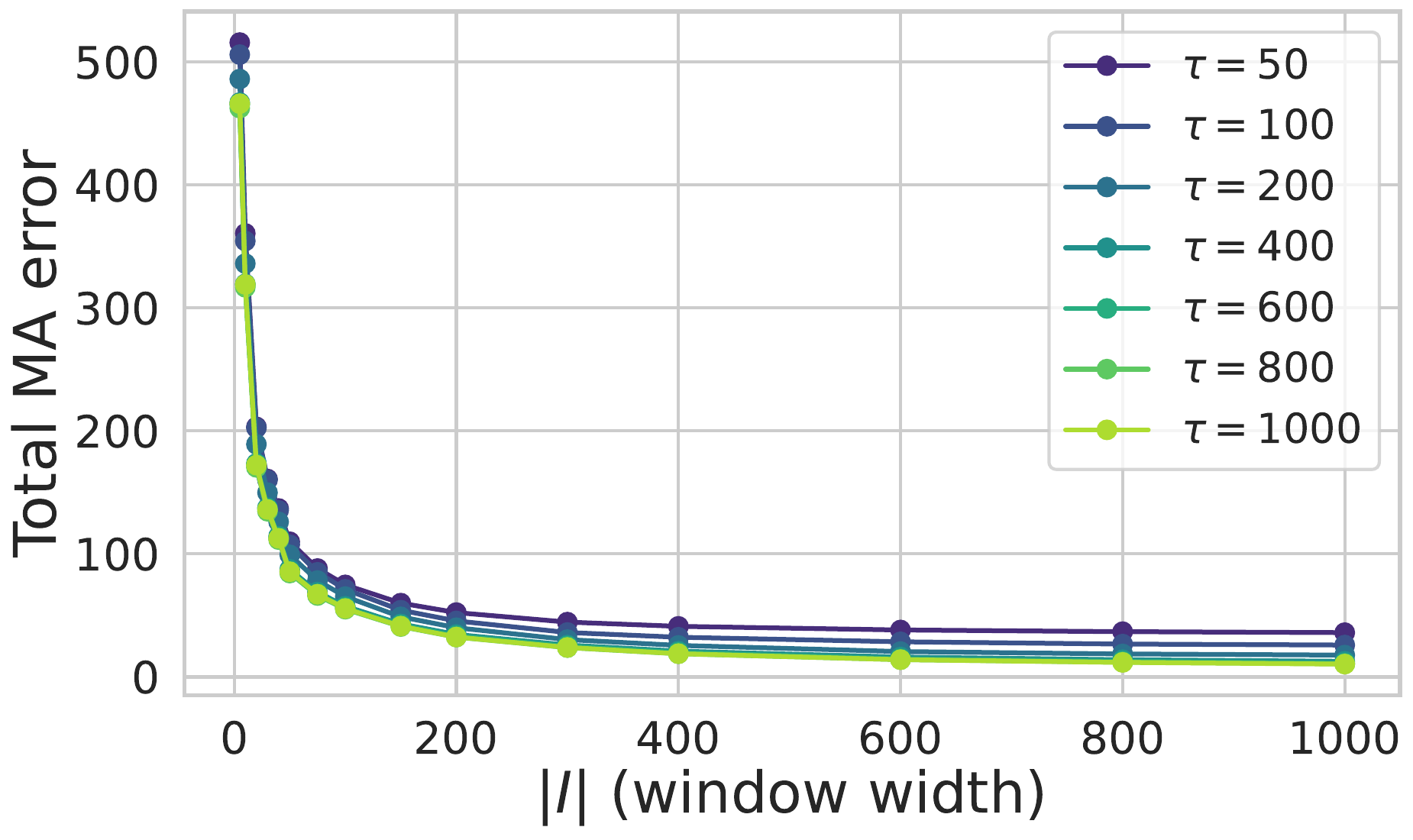}
        \caption{GEFCom2014-L}
        \label{fig:load_vary_gamma}
    \end{subfigure}
    \hfill
    \begin{subfigure}{0.47\textwidth}
        \centering
        \includegraphics[width=\linewidth]{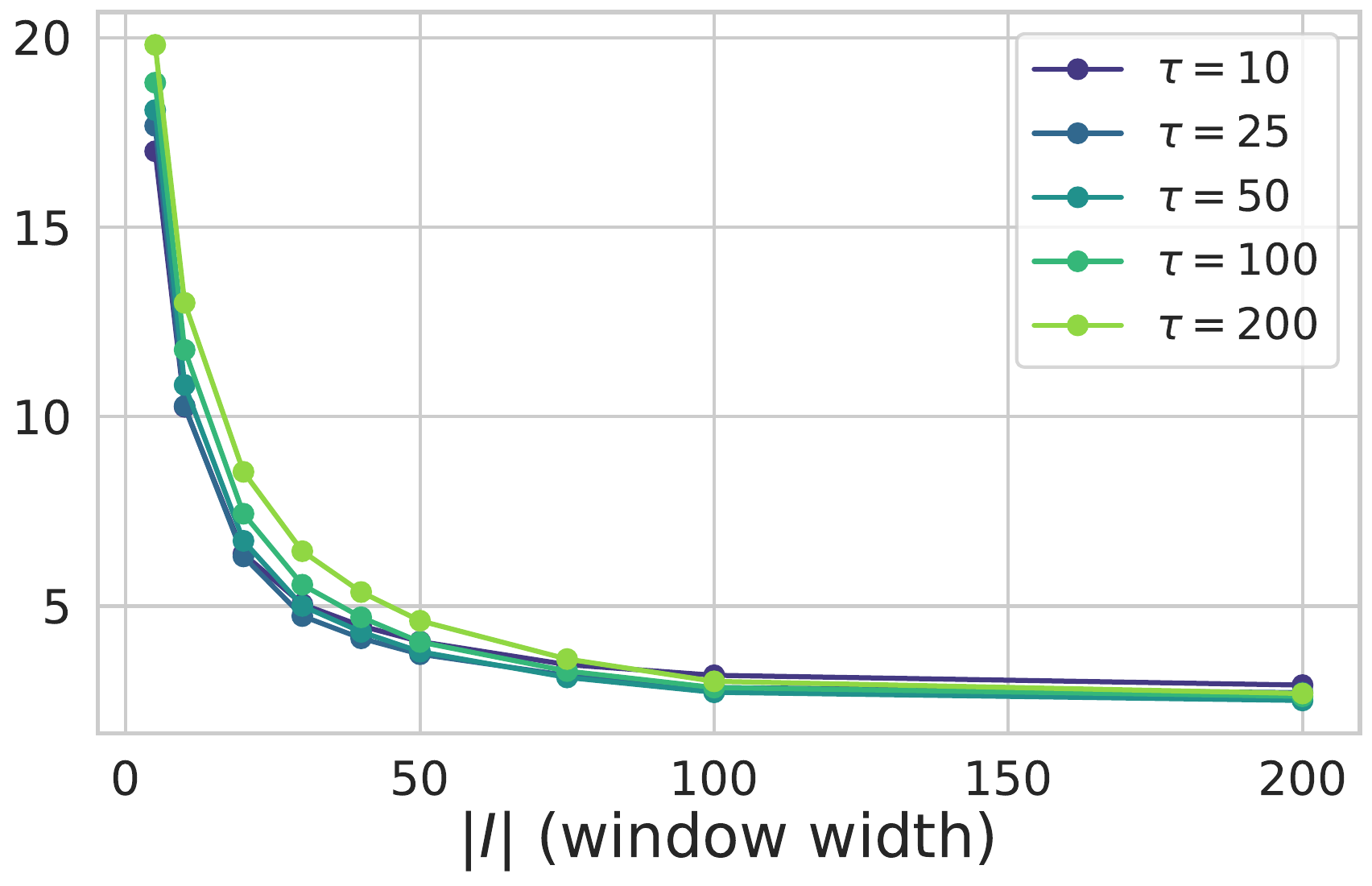}
        \caption{COMPAS}
        \label{fig:compas_vary_gamma}
    \end{subfigure}
   
    \caption{\textbf{Total multiaccuracy error for different $\tau$ with varying interval width $\boldsymbol{|I|}$,} (a) GEFCom2014-L and (b) COMPAS. This is the same setting as Figure \ref{fig:gefcom_adaptive} and Figure \ref{fig:compas_adaptive}. We vary the fixed width $\tau$ used for tuning MA+pred (locally adaptive) and the window width $|I|$ used for the moving average of errors and plot the total multiaccuracy error under the curve.}
    \label{fig:total_ma_vs_gamma}
\end{figure}

We now vary the fixed interval width $\tau$ used for tuning the locally adaptive MA+pred algorithm. This also results in different values of optimal $\gamma = 1/(2\tau)$. We evaluate the total multiaccuracy error for different choices of $\tau$ over windows of varying width $|I|$ in Figure \ref{fig:total_ma_vs_gamma}. Results show that the total error does not significantly change with different $\tau$ values and that the locally adaptive algorithm is robust to the choice of $\tau$.

\section{Simulated Examples}

\begin{figure}[thb!]
    \centering
    \begin{subfigure}[t]{0.95\linewidth}
        \includegraphics[width=\linewidth]{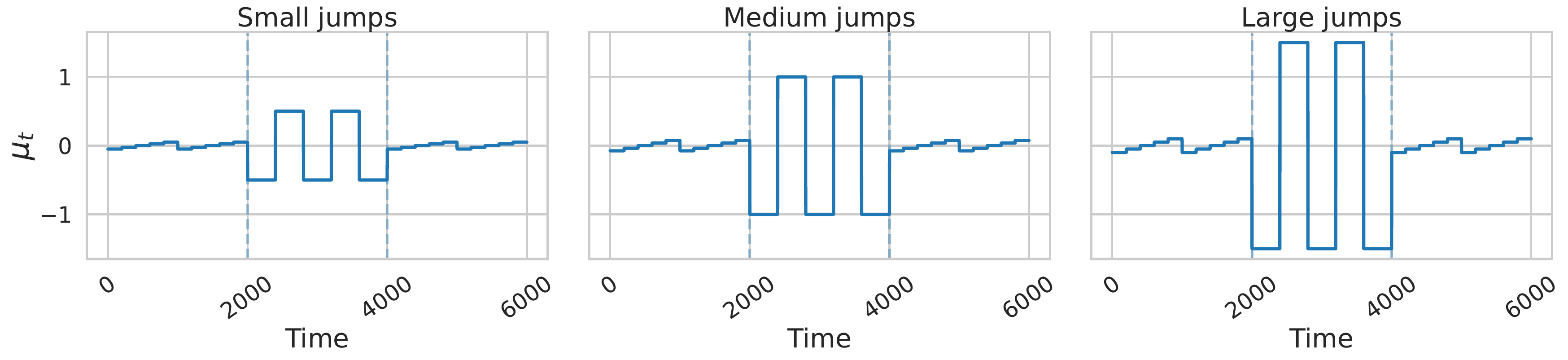}
        \caption{Trajectory for $\mu_t$ in the \textit{small} (left), \textit{medium} (mid), and \textit{large} (right) jump shift settings.}
        \label{fig:mu}
    \end{subfigure}
    \par\medskip
    \begin{subfigure}[t]{0.95\linewidth}
        \includegraphics[width=\linewidth]{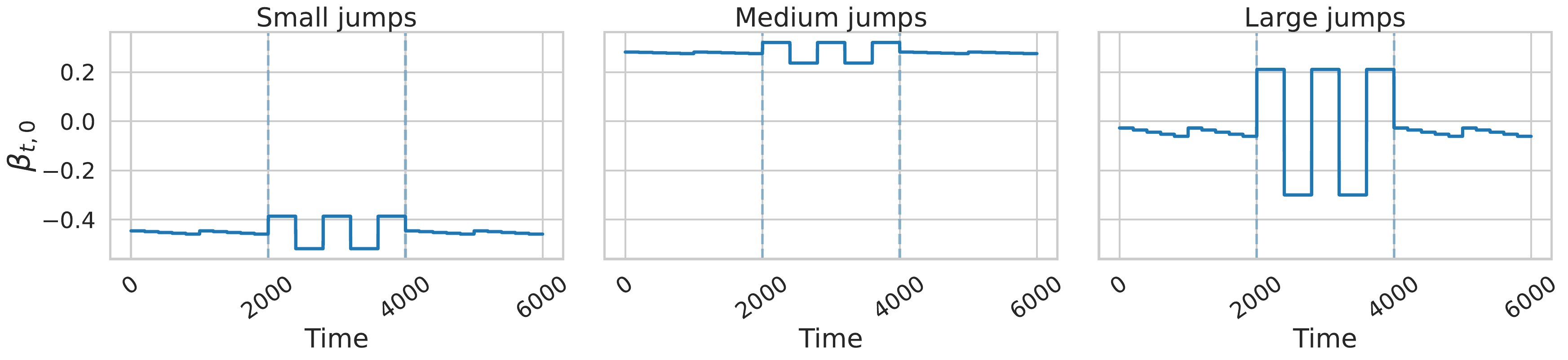}
        \caption{Trajectory for $\beta_{t,0}$ in the \textit{small} (left), \textit{medium} (mid), and \textit{large} (right) jump shift settings.}
        \label{fig:betas}
    \end{subfigure}
   
    \caption{\textbf{Trajectories for (a) $\boldsymbol{\mu_t}$ and (b) $\boldsymbol{\beta_{t,0}}$ in the different jump shift settings.} $\beta_{t,0}$ denotes the first coordinate of $\beta_t$. $\beta_{t,j}$ is an affine transformation of $\mu_t$ by construction. Dashed vertical lines denote the boundaries between the regime switches where the size of the distribution shift changes.}
    \label{fig:jump_shifts} 
\end{figure}

We consider a set of simulated examples where we can control the distribution shifts over time. We focus on a simple setting with a time-varying linear model
\[
Y_t \;=\; X_t^\top \beta_t + \varepsilon_t, 
\qquad \varepsilon_t \sim \mathcal{N}(0,\sigma^2),
\]
where the covariates $X_t$ are i.i.d.\ Gaussian,
\[
X_t \sim \mathcal{N}(0, I_d),
\]
and we specify the distribution shift entirely through the coefficients $\beta_t \in \mathbb{R}^d$. The initial $\beta_0 \sim \mathcal{N}\!\left(0, \tfrac{1}{d} I_d\right)$ and we set
\[
\beta_t \;=\; \beta_0 + \mu_t\, v,
\]
where $v \in \mathbb{R}^d$ is a unit direction vector sampled uniformly at random from the unit sphere and $(\mu_t)_{t=1}^T$ controls the magnitude of the shift along direction $v$. We consider \textit{jump} discontinuities in $\mu_t$ of varying sizes. Specifically, we divide the time horizon into three equally sized intervals and define $\mu_t$ to have small-amplitude jumps in the first and third intervals and large-amplitude jumps in the second interval. We construct three jump–shift datasets (\textit{small, medium}, and \textit{large}) by increasing the small-amplitude range and the large-amplitude range of
$\mu_t$ as
\begin{itemize}
    \item s\textit{mall }shift: $\mu_t$ oscillates between $[-0.05, 0.05]$ in the small-amplitude regime and between $[-0.5, 0.5]$ in the large-amplitude regime.
    \item \textit{medium} shift: $\mu_t$ oscillates between $[-0.075, 0.075]$ in the small-amplitude regime and between $[-1.0, 1.0]$ in the large-amplitude regime.
    \item \textit{large} shift: $\mu_t$ oscillates between $[-0.1, 0.1]$ in the small-amplitude regime and between $[-1.5, 1.5]$ in the large-amplitude regime.
\end{itemize}
We take $d=5$ in our experiments. Figure \ref{fig:jump_shifts} shows the final trajectories of $\mu_t$ and $\beta_{t,0}$, the first coordinate of $\beta_t$, in all three jump shift settings.

\begin{figure}[h!] 
    \centering
    \includegraphics[width=\linewidth]{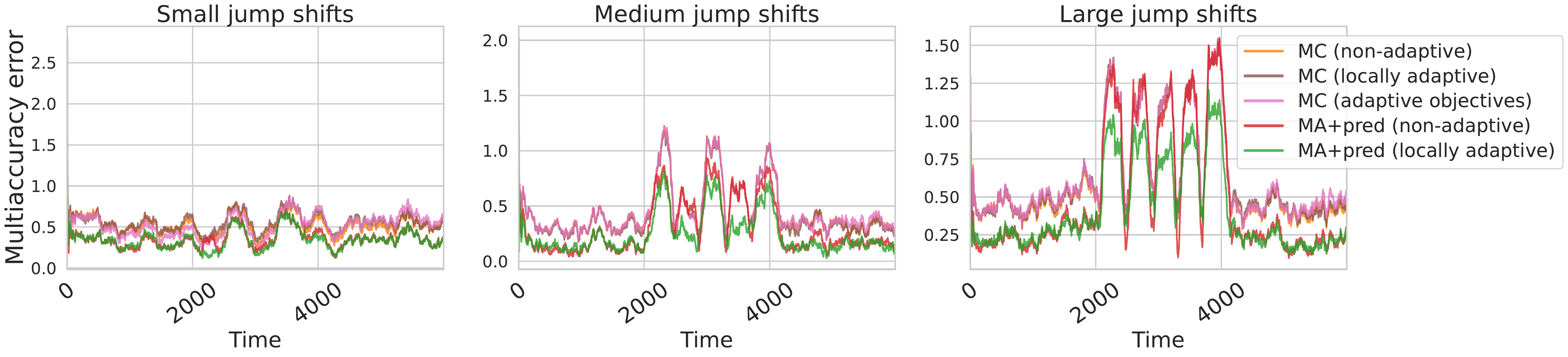} 
    \caption{\textbf{Local multiaccuracy error in different jump shift settings,} \textit{small} (left), \textit{medium} (mid), and \textit{large} (right) jump shifts.}
    \label{fig:jump_compare}
\end{figure}

We set the function class $\gF = \{f_j\}_{j=1}^d$ to the mappings onto each of the covariates given by $f_j(X) = X_j, \; j \in [d]$. Figure \ref{fig:jump_compare} shows the results comparing all algorithms across the three settings. While all algorithms reasonably adapt to the distribution shift in the small jump shift setting, MA+pred (locally adaptive) consistently outperforms all methods as the magnitude of shift increases. The difference is especially substantial in the large jump shift setting. Results from these examples show that the proposed locally adaptive algorithm is able to adapt to discontinuous and abrupt distribution shifts with better rates than existing methods.

